\documentclass[11pt]{article}
\usepackage{times}
\usepackage{psfrag,amsmath,amsbsy,amsfonts,amssymb,units}
\usepackage{graphicx,psfrag}
\usepackage{algorithm,algorithmic,mathtools,appendix}
\usepackage{color,cases}
\usepackage{tikz,natbib}
\usetikzlibrary{calc,shapes,fit}
\usepackage{fullpage}

\DeclareMathOperator{\errrank}{\epsilon_{\rank}}

\DeclareMathOperator{\ksvd}{k-svd}

\DeclareMathOperator{\vecform}{vec}
\DeclareMathOperator{\mat}{mat}
\DeclareMathOperator{\Proj}{Proj}
\DeclareMathOperator{\Projhat}{P\widehat{roj}}
\DeclareMathOperator{\Tr}{tr}

\DeclareMathOperator{\norm}{\Vert}
 \newcommand{\IGNORE}[1]{}

\def\nn{\nonumber}

\newcommand\E{\mathbb{E}}
\newcommand\R{\mathbb{R}}

\newcommand\thres{\text{Thres}}
\def\Pbb{\mathbb{P}}

\newcommand\h{\widehat}

\newcommand\tl{\widetilde}

\renewcommand\cite{\citep}

\def\tl{\tilde}

\newcommand\inner[1]{\ensuremath{\langle #1 \rangle}}

 \DeclareMathOperator*{\argmax}{arg\,max}

\DeclareMathOperator{\rank}{Rank}

\def\simiid{{\overset{i.i.d.}{\sim}}}

\def\tha{{\mbox{\tiny th}}}

\DeclareMathOperator{\Diag}{Diag}

 \def\0{{\bf 0}}

\DeclareMathOperator{\eqa}{\overset{(a)}{=}}
\DeclareMathOperator{\eqb}{\overset{(b)}{=}}
\DeclareMathOperator{\eqc}{\overset{(c)}{=}}
\DeclareMathOperator{\eqd}{\overset{(d)}{=}}

\def\viz{{viz.,\ \/}}

\def\nn{\nonumber}

\def\qed{\hfill\hbox{${\vcenter{\vbox{
    \hrule height 0.4pt\hbox{\vrule width 0.4pt height 6pt
    \kern5pt\vrule width 0.4pt}\hrule height 0.4pt}}}$}}

\def\tcr{\textcolor{red}}
\def\tcb{\textcolor{blue}}

\definecolor{myred}{rgb}{0.3,0.0,0.7}
\definecolor{dkg}{rgb}{0.1,0.7,0.2}
\definecolor{dkb}{rgb}{0.0,0.2,0.8}

 \def\hB{\hat{B}}

 \def\hG{\widehat{G}}

\def\hW{\widehat{W}}

\def\bfi{{\mathbf i}}
\def\bfj{{\mathbf j}}

\def\Tc{{\cal T}}

\def\Ebb{{\mathbb E}}
\def\Fbb{{\mathbb F}}

\def\Pbb{{\mathbb P}}

\def\Rbb{{\mathbb R}}

\newcommand{\bprfof}{\begin{proof_of}}
\newcommand{\eprfof}{\end{proof_of}}
\newcommand{\bprf}{\begin{myproof}}
\newcommand{\eprf}{\end{myproof}}
\newcommand{\bp}{\begin{psfrags}}
\newcommand{\ep}{\end{psfrags}}
\newcommand{\bl}{\begin{lemma}}
\newcommand{\el}{\end{lemma}}
\newcommand{\bt}{\begin{theorem}}
\newcommand{\et}{\end{theorem}}
\newcommand{\bc}{\begin{center}}
\newcommand{\ec}{\end{center}}
\newcommand{\bi}{\begin{itemize}}
\newcommand{\ei}{\end{itemize}}
\newcommand{\ben}{\begin{enumerate}}
\newcommand{\een}{\end{enumerate}}
\newcommand{\bd}{\begin{definition}}
\newcommand{\ed}{\end{definition}}
\def\beq{\begin{equation}}
\def\eeq{\end{equation}\noindent}
\def\beqn{\begin{eqnarray}}
\def\eeqn{\end{eqnarray} \noindent}
\def\beqnn{  \begin{eqnarray*}}
\def\eeqnn{\end{eqnarray*}  \noindent}
\def\bcase{  \begin{numcases}}
\def\ecase{\end{numcases}   \noindent}
\def\bsbcase{  \begin{subnumcases}}
\def\esbcase{\end{subnumcases}   \noindent}

\newtheorem{theorem}{Theorem}

\newtheorem{lemma}[theorem]{Lemma}

\newtheorem{proposition}{Proposition}

\newtheorem{definition}{Definition}

\newenvironment{myproof}{\noindent{\bf Proof:} \hspace*{1em}}{
    \hspace*{\fill} $\Box$ }
\newenvironment{proof_of}[1]{\noindent {\bf Proof of #1: }}{\hspace*{\fill} $\Box$ }

\newcommand{\matplottc}[1]{               
        \unitlength .45truein
        \begin{center}
        \includegraphics{#1.ps}
        \end{picture}
        \end{center}
}

\def\psfancypar#1#2{\begingroup\def\par{\endgraf\endgroup\lineskiplimit=0pt}
               \setbox2=\hbox{\large\sc #2}
               \newdimen\tmpht \tmpht \ht2 \advance\tmpht by \baselineskip
               \font\hhuge=Times-Bold at \tmpht
               \setbox1=\hbox{{\hhuge #1}}
               \count7=\tmpht \count8=\ht1
               \divide\count8 by 1000 \divide\count7 by \count8
               \tmpht=.001\tmpht\multiply\tmpht by \count7
               \font\hhuge=Times-Bold at \tmpht
               \setbox1=\hbox{{\hhuge #1}}
               \noindent
                \hangindent1.05\wd1
               \hangafter=-2 {\hskip-\hangindent
               \lower1\ht1\hbox{\raise1.0\ht2\copy1}%
                \kern-0\wd1}\copy2\lineskiplimit=-1000pt}

\def\Kout{\setbox1=\hbox{\Huge\bf K}\hbox to
1.05\wd1{\hspace{.05\wd1}
}}
\def\Sout{\setbox1=\hbox{\Huge\bf S}\hbox to 1.05\wd1{\hspace{.05\wd1}
}}

\newcommand{\torestate}[3]{%
\expandafter \def \csname BBRESTATE #2 \endcsname{#3}
\theoremstyle{plain}
\newtheorem{BBRESTATETHMNUM#2}[theorem]{#1}
\begin{BBRESTATETHMNUM#2}\label{#2}\csname BBRESTATE #2 \endcsname   \end{BBRESTATETHMNUM#2}
\newtheorem*{BBRESTATETHMNONNUM#2}{{#1}~\ref{#2}}
}

\newcommand{\restate}[1]{\begin{BBRESTATETHMNONNUM#1}[Restated] \csname BBRESTATE #1 \endcsname
\end{BBRESTATETHMNONNUM#1}}

\definecolor{blue1}{HTML}{0066FF}
\definecolor{lpurple}{cmyk}{.05,0.18,0,0}

\def\hW{\hat{W}}

\title{Learning  Mixed Membership Community Models \\in Social Tagging Networks through Tensor Methods}

\author{
  Anima Anandkumar\footnote{University of California, Irvine, Email: a.anandkumar@uci.edu}
  \and  
    Hanie Sedghi\footnote{University of Southern California, Email: hsedghi@usc.edu}
}

\begin{document}
\maketitle

\begin{abstract}Community detection in graphs has been extensively studied both in theory and in applications. However, detecting communities in hypergraphs is more challenging. In this paper, we propose a tensor decomposition approach for guaranteed learning of communities in a special class of hypergraphs   modeling social tagging systems or {\em folksonomies}. A folksonomy is a tripartite 3-uniform hypergraph consisting of (user, tag, resource) hyperedges. We posit a probabilistic {\em mixed membership} community model, and prove that the tensor method consistently learns the communities under efficient sample complexity and separation requirements.  
\end{abstract}


\section{Introduction}

Folksonomies or social tagging systems~\cite{chakraborty2012detecting}   have been hugely popular in recent years. These are tripartite networks consisting of users, resources and tags. The resources can vary according to the system.  For instance, in   Delicious, the   URLs are the resources, in Flickr, they are  the images, in LastFm, they are the music files, in MovieLens, they are the reviews, and so on. The collaborative annotation of these resources by users  with descriptive keywords, enables faster search and retrieval~\cite{chakraborty2013clustering}.

The role of community detection in folksonomies cannot be overstated. Online social tagging systems are growing rapidly and it is important to group the nodes (i.e. users, resources and tags) for scalable operations in a number of applications such as personalized search~\cite{xu2008exploring},  resource and friend recommendations~\cite{konstas2009social}, and so on. Moreover, learning communities  can provide an understanding of community formation behavior of humans,  and the role of communities in human interaction and collaboration in online systems.

Folksonomies are special instances of hypergraphs. A folksonomy is a tripartite $3$-uniform hypergraph consisting of hyperedges between users, resources and tags. Scalable community detection in hypergraphs is in general challenging, and most previous works are limited to pure membership models, where a node belongs to at most one group. This is highly unrealistic since users have multiple interests, and the tags and resources have multiple contexts or topics. A few works which do consider overlapping communities in folksonomies are heuristic without any guarantees and do not incorporate any statistical modeling  (see Section~\ref{sec:related} for details).

In this paper, we propose a novel probabilistic approach for modeling folksonomies, and  propose a guaranteed approach for detecting overlapping communities in them. A naive model for folksnomies would result in  a large number of   model parameters, and make learning intractable. Here we present a more scalable approach where  realistic conditional independence constraints are imposed, leading to scalable modeling and tractable learning.

Our model is  a hypergraph extension of the popular {\em mixed membership stochastic blockmodel} (MMSB), introduced by Airoldi et. al~\cite{ABFX08}. We impose additional conditional independence constraints, which are natural for social tagging systems. We term our model as {\em mixed membership stochastic folksonomy} (MMSF). When hypergraphs are generated from such a class of MMSFs, we show that the hyper-edges can be much {\em more informative} about the underlying communities, than in the graph setting. Intuitively, this is because the hyper-edges represent multiple {\em views} of the the hidden communities. In this paper, we show that these properties can be exploited for learning via spectral approaches.

\subsection{Summary of Results}

We develop a practically relevant mixed membership hypergraph model and propose novel methods to learn them with guarantees. 
We posit a probabilistic model for generation of hyper-edges $\{r,u,t\}$ between resources $r$, users $u$ and tags $t$. We impose natural conditional independence assumptions that conditioned on the community memberships of individual nodes, the hyperedge generations are independent. In addition, we assume that the users select tags for a given resource, based on the context in which the resource is accessed. For instance, consider the resource as a paper that falls both in theoretical and applied machine learning, as shown in Figure~\ref{fig:mmsf}.  If a user accesses the resource under the context of theory, he/she uses tags that are indicative of theory. Note that we allow the users and tags to  be in multiple communities; however, the actual realization of an hyper-edge depends only on the context in which the resource was accessed. Depending on what kind of user is tagging the paper, the likelihood of choosing various tags such as application, latent variable model etc changes. The conditional independence assumption states that once a user accesses the paper in certain context (e.g. looking for applications), the probability of using tags in a category (e.g. applications, experiments) only depends on that context. There are many other such examples.  For example, a movie can be a drama about a political figure. A person who is mostly into politics will watch this movie in the context of politics and use political tags (for example name of the person, specific political events that where illustrated in the movie), while a person who is more into drama genre will use drama to tag the movie.

While community models on general hypergraphs is NP hard, our setting is geared towards the setting of folksonomies with users, resources and tags, and the assumptions we make naturally hold in this setting. Importantly, we allow for general distributions for mixed community memberships. The earlier work by~\citet{AnandkumarEtal:community12} on MMSB models on graphs is limited to the Dirichlet distribution. Note that the Dirichlet assumption for community memberships can be limiting and cannot model general correlations in memberships. Without the Dirichlet assumption, the earlier techniques, when applied directly, would yield tensors in the Tucker form, which do not possess a unique decomposition and thus, the communities cannot be learnt from the tensor forms.  In addition, our moment forms are different since it is the hypergraph setting and conditional independence assumptions are different. Thus, earlier work on MMSB cannot be directly applied here.

In addition,  we impose weak assumptions on the distribution of the community memberships. This is required since the memberships are in general not identifiable when they are mixed. While the original MMSB model~\cite{ABFX08} assumes that the communities are drawn from a Dirichlet distribution, here, we do not require such a strong parametric assumption.    Here, we impose a weak assumption that   a certain fraction of resource nodes are ``pure'' and belong to a single community. This is reasonable to expect in practice.
We establish that the communities are identifiable under these natural assumptions, and can be learnt efficiently using spectral approaches.

Here, we propose a novel algorithm to detect pure nodes belonging to a single community. The presence of pure nodes is natural to expect in practice and does not require the Dirichlet assumption. 
Our method consists of two main routines. First, we design a simple rank test to identify pure resource nodes.
The algorithm involves first projecting hyperedges to subspace of top-k eigenvectors. It then involves performing rank test on the matricization of connectivity vectors of each resource node, where rows correspond to users and columns correspond to tags.     
We can then exploit these detected pure nodes to form tensors that can be decomposed efficiently to yield the communities for all the nodes (and not just the pure nodes).
 We prove that our proposed method correctly recovers the parameters of the MMSF model when exact moments are input.    This two stage algorithm is expected to have much wider applicability than the MMSB model which is limited to the Dirichlet distribution. For this general model, we show a tight sample complexity that $n>k^3$ can recover the communities. 

For the first step, we construct a matrix for each resource node, consisting of its edges to users and tags. We show that this matrix is rank-$1$ in expectation (over the hyperedges) for a pure resource node. This property enables us to identify such pure nodes. We then construct a $3$-star count tensor using these estimated pure resource nodes. We count the pure resource nodes, which are common to  triplets of (user,tag) tuples to form the tensor. We show that in expectation this tensor has a CP decomposition form, and requiring this decomposition yields the community memberships after some simple post-processing steps.

We then carefully analyze the perturbation bounds under empirical moments, and show that the communities can be accurately recovered under some natural assumptions.  The perturbation analysis for this step is novel since it requires analyzing the effect of standard spectral perturbations on matricization and the subsequent rank test. We use subexponential Hanson Wright inequalities to obtain tight guarantees for this step. These assumptions determine how the number of nodes $n$ is related to the number of communities $k$,  and a lower bound on the separation $p-q$, where $p$ denotes the connectivity within the same community, while $q$ denotes the connectivity across different communities. Such requirements have been imposed before in the graph setting, for stochastic block models~\citep{ChenSanghaviXu} and mixed membership models~\citep{AnandkumarEtal:community12}. Here, we show that for MMSF, the requirement is stronger, since intuitively,  we require concentration on a hypergraph instead of a graph.  We employ sub-exponential forms of Hanson Wright's inequality to get tight bounds in the sparse regime, where the connectivity probabilities $p,q$ are small. Thus, we obtain efficient guarantees for recovering mixed membership communities from social tagging networks.

We establish that for the success of rank test,  if $p \simeq q$, we need the network size to scale as $n= \tilde{\Omega}\left( k^3\right)$ (when the correlation matrix of community membership distribution is   well-conditioned). For the case where $q < p/k$, we   require $n=\tilde{\Omega}\left( k^2\right)$. This is intuitive as  the role of $q$ is to make the different community components non-orthogonal for the rank test, i.e., $q$ acts as noise. Therefore, a smaller $q$ results in better guarantees.
For the success of tensor decomposition method, we require $n= \tilde{\Omega}\left( k^3\right)$, when $p, q$ are constants, in the well-conditioned setting. Note that in comparison, for learning mixed membership stochastic block model graphs, we require $n = \tilde{\Omega}\left( k^2\right)$, from~\citet{AnandkumarEtal:community12}, which is lower sample complexity. This is because we need to learn more number of parameters in the hypergraph setting. Moreover, for sparse graphs, the parameters  $p, q$   decay with $n$, and we also handle this setting, and provide the precise bounds in Section~\ref{sec:special}.

\subsection{Related Work} \label{sec:related}

%


There is an extensive body of work for community detection in graphs. Popular methods with guarantees include spectral clustering~\cite{McSherry01} and convex optimization~\cite{ChenSanghaviXu}. For a detailed survey, see~\cite{AnandkumarEtal:community12}.  However, these methods cannot handle mixed membership models, where a node can belong to more than one community.

Our algorithm is based on the tensor decomposition approach of~\cite{AnandkumarEtal:community12COLT} for pairwise MMSB model in graphs. The method has been implemented for many real-world datasets and has shown significant improvement in running times and accuracy over the state of art stochastic variational techniques~\cite{AnandkumarEtal:communityimplementation13}. The tensor consists  of third order moments in the form of counts of $3$-star subgraphs, i.e., a star subgraph
consisting of three leaves, for each triplet of leaves.
The MMSB model assumes a Dirichlet distribution for community memberships, and in this case, a modified $3$-star count tensor is used. It is shown that this tensor has a {\em CP}-decomposition form, and the components of the decomposition can be used to learn the parameters of the MMSB model. However, this method cannot be extended easily to  general distributions, beyond the Dirichlet assumption, since for general distributions, the $3$-star count tensor only has   a Tucker decomposition form, and not a CP form. In general, the model parameters are not  identifiable from a Tucker form. Thus, in graphs, mixed membership models cannot be easily learnt when general distributions (beyond the Dirichlet distribution) for mixed memberships are assumed. In this paper, we show that in the hypergraph setting, more general distributions of community memberships can be learnt, when certain conditional independence relationships are assumed for hyper-edge generation.

Another limitation of the MMSB model is that due to the Dirichlet assumption, only normalized community memberships can be incorporated. However, in this case, the mixed nodes (i.e. those belonging to more than one community) are less densely connected than the pure nodes, as pointed out by~\cite{yang2013overlapping}.  In contrast, in our paper, we can handle   un-normalized community memberships vectors (in  a weighted graph), since we do not make the Dirichlet assumption, and thus, this limitation is not present. However, for simplicity, we present the results in the normalized setting.


Scalable community detection in hypergraphs is in general challenging and most previous works are limited to pure membership models, where a node belongs to at most one group~\cite{brinkmeier2007communities,lin2009metafac,murata2010detecting,neubauer2009towards,vazquez2009finding}. Clustering in multipartite hypergraphs can be seen as extensions of the {\em co-clustering} of matrices, where rows and columns are simultaneously clustered. In~\cite{jegelka2009approximation}, extensions of co-clustering to the tensor setting is considered. However, this setting can only handle pure communities, where a node belongs to at most one community. A few works which do consider mixed communities in hypergraphs  are heuristic without any guarantees, and do not incorporate any statistical modeling~\cite{wang2010discovering,chakraborty2012detecting,papadopoulos2010graph}. They mostly use modularity based scores without providing any guarantees. In this paper, we present the first guaranteed method for learning communities in mixed membership hypergraphs.
%
%
%
%

\section{Mixed Membership  Model for Folksonomies}\label{sec:model}


\begin{figure}[t]
\bc
\begin{tikzpicture}

\node [anchor=west, outer sep=-3pt](Comm) at (4,1) {\em \footnotesize Communities:};
\node [anchor=west, outer sep=-3pt](T-ML) at (4,0.5) {\footnotesize 1) T-ML: Theoretical Machine Learning};
\node [anchor=west, outer sep=-3pt](A-ML) at (4,0) {\footnotesize 2) A-ML: Applied Machine Learning};
\node [draw, black, fit = {(Comm) (T-ML) (A-ML)}] {};  

\node [outer sep=-10pt, label=left:{\begin{tabular}{l} \small\footnotesize 90\% T-ML \\ \footnotesize 10\% A-ML \end{tabular}}](user1) at (-3,-2) {\includegraphics[width=.05\textwidth]{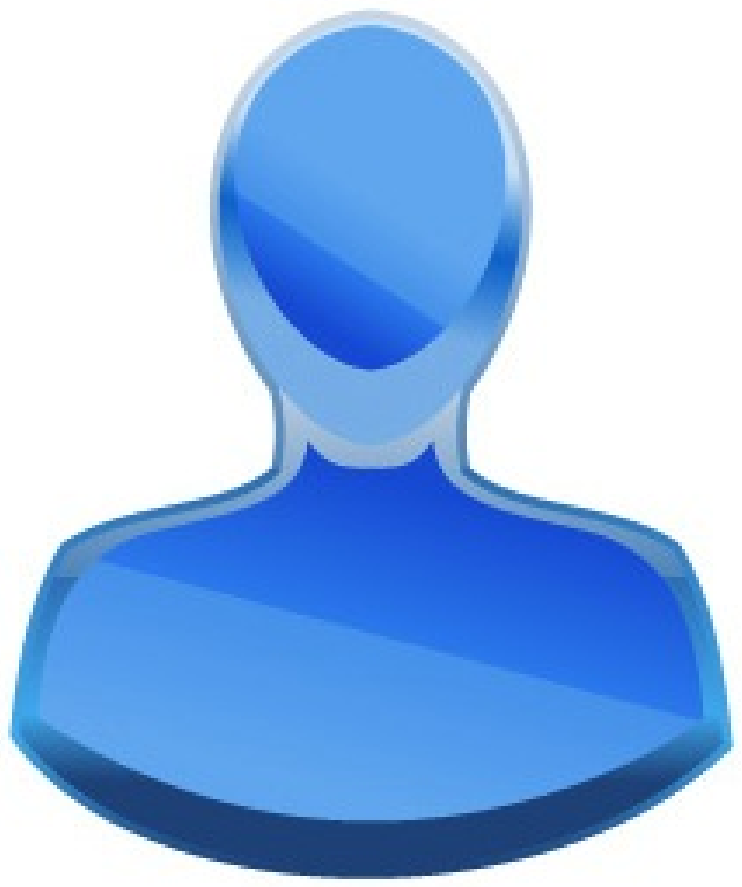}};
\node [draw, dashed, thick, rounded corners, blue!80!white, fit = {(user1) ($(user1.west)-(18mm,0)$) ($(user1.east)+(2mm,0)$) ($(user1.north)+(0,2mm)$) ($(user1.south)-(0,2mm)$)}, label=north:{\em User 1} ] {};

\node [outer sep=-10pt, label=left:{\begin{tabular}{l} \small\footnotesize 10\% T-ML \\ \footnotesize 90\% A-ML \end{tabular}}](user2) at (-2,-3.5) {\includegraphics[width=.05\textwidth]{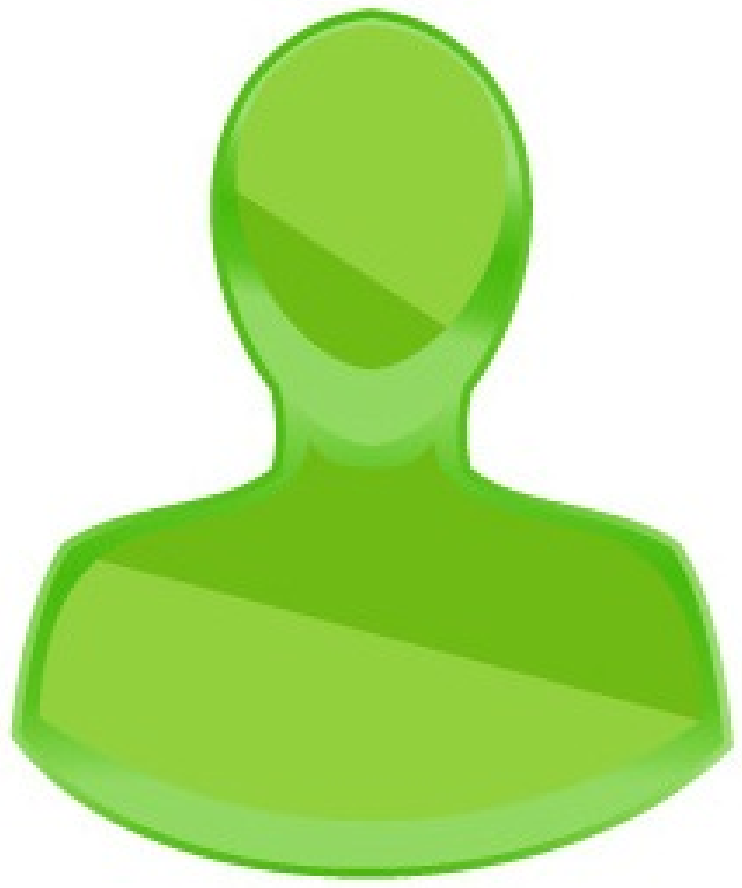}};
\node [draw, dashed, thick, rounded corners, green!80!black, fit = {(user2) ($(user2.west)-(18mm,0)$) ($(user2.east)+(2mm,0)$) ($(user2.north)+(0,2mm)$) ($(user2.south)-(0,2mm)$)}, label=south:{\em User 2} ] {};

\node [outer sep=-10pt, label=left:{\begin{tabular}{l} \small\footnotesize 55\% T-ML \\ \footnotesize 45\% A-ML \end{tabular}}](article) at (0,0) {\includegraphics[width=.15\textwidth]{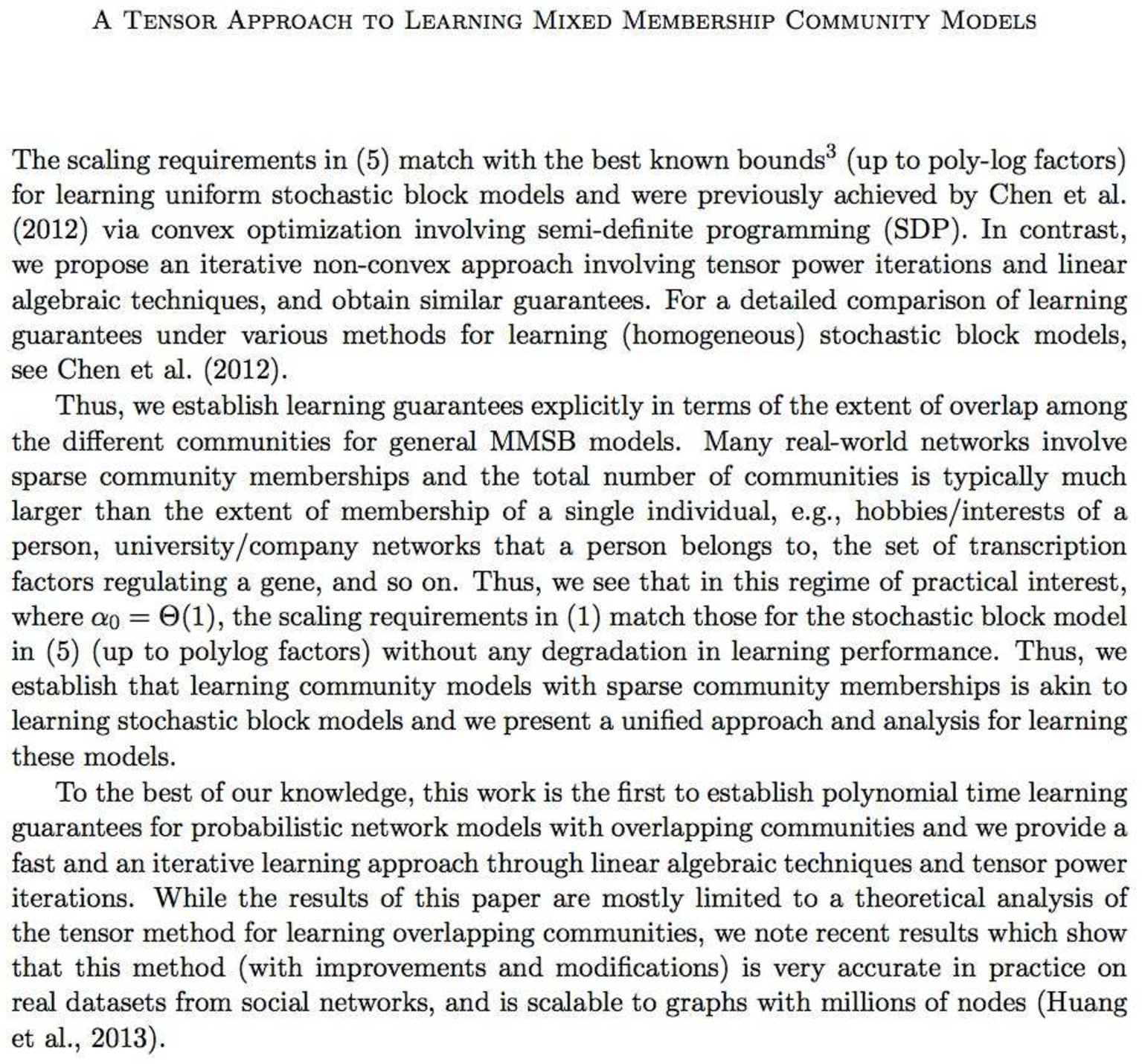}};
\node [draw, dashed, thick,  rounded corners, violet, fit = {(article) ($(article.west)-(18mm,0)$) ($(article.north)+(0,2mm)$) ($(article.south)-(0,2mm)$)}, label=north:{\em Resource 1} ] {};  

\node (rect) at (-1,-2) [minimum width=0.2cm,fill=blue!80!white,name=joint1] {};
\node (rect) at (0.5,-2) [minimum width=0.2cm,fill=green!80!black,name=joint2] {};

\node [anchor=west,name=app] at (1.5,-2.2) {Applications ({\footnotesize 20\% T-ML, 80\% A-ML})};
\node [anchor=west,name=LVMs] at (1.5,-2.8) {Latent Variable Models ({\footnotesize 35\% T-ML, 65\% A-ML})};
\node [anchor=west,name=theory] at (1.5,-3.4) {Theoretical Guarantees ({\footnotesize 95\% T-ML, 5\% A-ML})};
\node [anchor=west,name=rest] at (1.5,-4) {$\dotsb$};
\node [draw, dashed, thick, rounded corners, violet, fit = {(app) (theory) (LVMs) (rest)}, label=north:{\em Tags} ] {};  

\draw [-,color=green!80!black,thick] (app.west) to (joint2); 
\draw [-,color=green!80!black,thick] (LVMs.west) to (joint2);
\draw [-,color=green!80!black,thick] (user2) to (joint2);
\draw [-,color=green!80!black,thick] (article) to (joint2);
\draw [-,color=blue!80!white,thick] (theory.west) to (joint1); 
\draw [-,color=blue!80!white,thick] (LVMs.west) to (joint1);
\draw [-,color=blue!80!white,thick] (user1.east) to (joint1);
\draw [-,color=blue!80!white,thick] (article) to (joint1);

\end{tikzpicture}
\ec \label{fig:mmsf}
\vspace{-0.2in}
\caption{Overview of MMSF model for an example of machine learning articles (resources) tagged by users. One article (resource) and the corresponding tags by two users are shown. Two communites of Theoretical machine learning and Applied machine learning are assumed. The mixed community membership of resources, users and tags are also shown.}
\end{figure}

\paragraph{Setup: }We consider folksonomies  modeled as tripartite $3$-uniform hypergraphs over three sets of nodes, \viz
set of users $U$, set of tags $T$ and set of resources $R$. An hyperedge $\{u,t, r\}$ occurs when user $u$  tags resource $r$ with tag $t$. For convenience, we will consider a {\em matricized} version of the $\{0,1\}$ hyper-adjacency tensor, denoted by   $\h{G} \in \lbrace 0,1 \rbrace^{  |U|\cdot |T| \times |R|}$, which indicates the presence of hyper-edges. The reason behind considering matricization along the resource {\em mode} will soon become clear.
 We use the notation $\h{G}(\{u,t\}, r ) $ to denote the entry corresponding to the hyper-edge $\{u,t, r\}$, and $\hG(\{U,T\}, r)$ to denote the column vector corresponding to the set of hyper-edges $\{U,T,r\}$. 

We consider models with $k$ underlying (hidden) communities and let $[k] :=
\{1,2,\dotsc,k\}$. For node $i$, let $\pi_i\in \R^k$ denote its \emph{community membership vector}, i.e., the vector is supported on the communities to which the node belongs.
Define
$\Pi_U:=[\pi_i : i \in U] \in \R^{k \times |U|}$ denote the  set of column
vectors  denoting the community memberships of users in $U$, and similarly define $\Pi_T$ and $\Pi_R$. Let $\Pi:= [\pi_i : i \in U\cup T\cup R] $.

We now provide a statistical model to explain the presence of hyper-edges $ \{u,t,r\}$ among users, tags and resources through the community memberships.  We consider a mixed memberships model, where
there are multiple communities for users, tags and resources.
Intuitively, users belonging to certain groups (i.e. interested in certain topics) will tend to select resources mainly comprised of those topics. The tags employed by the users are dependent on the {\em contextual} category of the resource selected by the user. This intuition is formalized under our proposed statistical model below.

Let $z_{u \rightarrow \{t,r\}}\in \R^k$ be a coordinate basis vector which  denotes the community membership of user $u$ when posting tag $t$ and resource $r$, and similarly let $z_{r \rightarrow \{u,t\}}$, $z_{t \rightarrow \{u,r\}}$ denote the memberships of resource $r$ and tag $t$ when participating in the hyperedge $ \{u,t,r\}$.

Let   $P \in \Rbb^{k\times k}$ be the community connectivity matrix, where $P_{i,j}$ denotes the probability that a user in community $i$ selects a resource in community $j$. Similarly,  let $\tl{P} \in \Rbb^{k \times k}$ denote a matrix such that each entry $\tl{P}_{i,j}$ denotes the probability that a tag in community $i$ is associated with  resource in community $j$.


The proposed mixed membership stochastic folksonomy (MMSF)   is as follows:

\bi \item For each node in $i \in U\cup T\cup R$, draw its community membership vector $\pi_i \in \R^k$, i.i.d. from some distribution $f_{\pi}$.

\item For each triplet $\{u , t, r\}$, draw coordinate basis vectors
 $z_{u \rightarrow \{t,r\}} \sim \mbox{Multinomial}(\pi_u)$,
  $z_{t \rightarrow \{u,r\}} \sim \mbox{Multinomial}(\pi_t)$ and  $z_{r \rightarrow \{u,t\}} \sim \mbox{Multinomial}(\pi_r)$ in a conditionally independent manner, given $\Pi$.

\item Draw random variables
 \begin{align}
  \nn\h{B}_{r\rightarrow u;t} &\sim \mbox{ Bernoulli}(  z_{u \rightarrow \{t,r\}}^\top  P  z_{r \rightarrow \{u,t\}})\\ \h{B}_{r\rightarrow t;u} &\sim\mbox{ Bernoulli}(  z_{t \rightarrow \{u,r\}}^\top  \tilde{P} z_{r \rightarrow \{u,t\}}).\label{eqn:bern}\end{align} The presence of  hyper-edge $G(\{u,t\},r)$ is given by the product
 \beq \h{G}(\{u,t\},r) =\h{B}_{r\rightarrow u;t} \cdot \h{B}_{r\rightarrow t;u}.\label{eqn:hyper}\eeq

 \ei

The use of variables $z_{u \rightarrow\{t,r\}}$, $z_{t \rightarrow\{u,r\}}$ and $z_{r \rightarrow\{u,t\}}$ allows for {\em context-dependent} selection of group memberships as in the MMSB model. 
 Given a resource and its context, a user may choose to access the resource, and probability of using a tag on a resource depends on context of the tag and the resource. Given the context of user, tag and the resource, these two events are independent. In order to have a hyper-edge, we need both events to happen and this explains Eqn. \eqref{eqn:hyper}.  

Ours is a resource centric model, where a resource can be regarded as comprising of many {\em topics} or communities. Which  tags get associated with the resource is dependent on the context of the resource $z_{r\rightarrow \{u,t\}}$ and the tag $z_{t\rightarrow \{u,r\}}$ and similarly, which   user selects a resource is dependent on the context of the user ($z_{u\rightarrow \{t,r\}}$)  and the resource
$z_{r\rightarrow \{u,t\}}$. The hyper-edges are drawn according to \eqref{eqn:hyper} and thus, matricization along the resource mode is convenient for analysis. Our model is resource centric and not user centric.  The intuition is that the tags associated with a resource are dependent on the context that the resource is being accessed and the likelihood of the user accessing a resource is dependent on his/her current group and the context of the resource. Figure~\ref{fig:mmsf} provides an instance of a hypergraph where the resource is a paper and communities consist of theoretical and applied machine learning.

Unlike the pairwise MMSB model~\cite{ABFX08}, where the edges are conditionally independent given the community memberships, in the proposed MMSF model, the edges $\h{B}_{r\rightarrow t;u}$ and $\h{B}_{r \rightarrow u;t}$ contained in the hyperedge $\{u,t, r\}$ are {\em not} conditionally independent given the community memberships, since they are selected based on the common context $z_{r\rightarrow \{u,t\}}$ of the resource $r$. Thus, the MMSF model is capturing dependencies beyond the pairwise MMSB model. At the same time, the MMSF model has conditionally independent hyperedges given the community memberships, which leads to tractable learning.

We do {\em not} take the approach of modeling hyperedges  directly, i.e., through a community connectivity tensor in $\tilde{P}\in\R^{k\times k \times k }$, where $\tilde{P}_{a,b,c}$   would give the probability that a user in community $a$ would have an hyperedge with resource $b$ and tag $c$. This would lead to $k^3$ unknown parameters, while our model has only $k^2$ unknown parameters. Moreover, if the user at a certain point is interested in some topic (i.e. draws $z_{u \rightarrow\{t,r\}}$ in some community), then he looks for resources and tags having significant membership in that topic (modeled through draws of $z_{t \rightarrow\{u,r\}}$ and $z_{r \rightarrow\{u,t\}}$) and this will generate the hyper-edge $u \rightarrow \{t,r\}$. 



We assume that the community vectors are drawn i.i.d. from a general unknown distribution: for $i\in [n]$,  $\pi_i \simiid  f_{\pi}(\cdot)$, supported on the $k-1$-dimensional simplex $\Delta^{k-1}$\[ \Delta^{k-1}:=\{\pi \in \Rbb^k, \pi(i)\in [0,1], \sum_i \pi(i)=1\}.\] The performance of our learning algorithms will depend on the distribution of $\pi$. In particular, we assume that with probability $\rho$, a realization of $\pi$ is  a  coordinate basis vector,  and thus, about $\rho$ fraction of the nodes in the network are {\em pure}, i.e. they belong mostly to a single community. In this paper, we investigate how the tractability of learning the communities depends on $\rho$.

\section{Proposed Method}


\paragraph{Notation: }For a matrix $M$, if $M = UD V^\top$ is the SVD of $M$, let $\ksvd(M):= U \tl{D} V^\top$ denote the $k$-rank SVD of $M$, where $\tl{D}$ is limited to top-$k$ singular values of $M$. A matrix $A\in \Rbb^{p \times q}$ is stacked as a vector $a \in \Rbb^{pq}$ by the $\vecform(\cdot)$ operator,
\begin{align*}
a = \vecform(A) \Leftrightarrow a \bigl(   (i_{1}-1)q + i_2)  \bigr) = A(i_1,i_2).
\end{align*} The reverse matricization operation is denoted by $\mat(\cdot)$, i.e. above $A=\mat(a)$. Let $A*B$ denote the Hadamard or entry-wise product. Let $\ksvd(M)$ of a matrix $M$ denote its restriction to top-$k$ singular values, i.e. if $M=U\Lambda V^\top$, $\ksvd(M) = U_k \Lambda_k V_k^\top$, which denote the restriction of the subspaces and the singular values to the top-$k$ ones.

In this paper, we consider the problem of learning the community vectors $\pi_i$, for $i\in [n]$, given a realization of the (matricized) hyper-adjacency matrix $G\in \R^{|R| \times |U|\cdot |T|}$. We will  employ a clustering-based approach on the hyper-adjacency matrix, but employ a different clustering criterion than the usual distance based clustering. our method is shown in Algorithm~\ref{algo:main}. 

Our method relies on finding {\it pure} resource nodes and using them to find communities for the resource, tag and users. A pure resource node is a node that is mainly corresponding to one hidden community. Therefore, finding that node paves the way for finding resource communities. In addition, since this is a resource-centric model, looking at the subset of hyper graph with pure resources, all tags and all users, suffices to find the communities for users and tags as well. Since we assume knowledge of community connectivity matrices, we can learn community memberships for mixed resource nodes as well. We now provide the details of our proposed method. 

 
\begin{figure}
\centering\bp
\psfrag{X}[c]{\hspace{0.1cm}\tcr{
 $ \tl{R}$}}\psfrag{j}[l]{\hspace{-.2cm}\scriptsize \tcb{$\lbrace u_i, t_i\rbrace$}}
\includegraphics[width=2in]{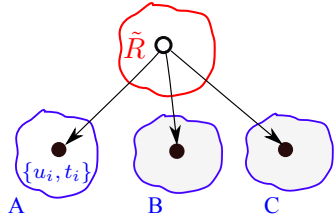}
\ep
\caption{Our moment-based learning algorithm uses 3-star count tensor from set $X$ to sets $A, B, C$.}
\label{fig:partition}
\end{figure}

\paragraph{Projection matrix: }
We partition the resource set $R$ into two parts $X$ and $Y$ to avoid dependency issues between the projection matrix and the projected vectors, and this is standard for analysis of spectral clustering. Now let $\ksvd(\hG(\{U,T\},Y))= M_k \Lambda_k V_k^\top$ and we employ  $\Projhat := M_k M_k^\top$ as the projection matrix. We project the vectors $\hG(\{U,T\},x)$ for $x\in X$ using this projection matrix.

\paragraph{Rank test on projected vectors: }In the usual spectral clustering method, once we have projected vectors $\Projhat\hG(\{U,T\},x)\in \R^{|U|\cdot |T|}$, any distance based clustering can be employed to classify the vectors into different (pure) communities. However, when mixed membership nodes are present, this method fails. We propose an alternative method which considers a rank test on the (matricized form of) the projected vectors. Specifically consider the matricized form
$ \mat(\Projhat\hG(\{U,T\},x)\in \R^{|U| \times |T|}$ and check whether
\[\sigma_1(\mat(\Projhat\hG(\{U,T\},x)))>\tau_1 \quad \text{and} \quad  \sigma_2(\mat(\Projhat\hG(\{U,T\},x)))< \tau_2 \] and if so, declare the node $x \in X$ as a {\em pure} node. Interchange roles of $X$ and $Y$ and similarly find pure nodes in $Y$.

\paragraph{Learning using estimated pure nodes: }Once the pure nodes in resource set  $R$ are found, we can employ the tensor decomposition method, proposed in~\cite{AnandkumarEtal:community12}, for learning the mixed membership communities of all the nodes. The pure nodes are employed to obtain averaged $3$-star subgraph counts. Partition $\{U,T\}$ into three sets $A, B, C$ as shown in Figure~\ref{fig:partition}. The $3$-star subgraph count  is defined as
\beq\label{eqn:threestardef} \h{\Tc}_{\tl{R}\rightarrow A,B,C} :=\frac{1}{|\tl{R}|} \sum_{r \in \tl{R}} \h{G}(r,A)^\top \otimes \h{G}(r,B)^\top \otimes \h{G}(r,C)^\top,
\eeq  where $\tl{R}$ denotes the set of pure resource nodes. The method is explained in Appendix~\ref{app:tensor}.


\paragraph{Reconstruction after power method: }Since we do not have access to the exact moments we need to do additional processing:  the estimated community membership vectors are then subject to thresholding so that the weak values are set to zero. This modification makes our reconstruction strong as we are considering sparse community memberships. Also note that assuming knowledge of community connectivity matrices, we can learn community memberships for mixed resource nodes as well. This is shown in Algorithm~\ref{alg:tensordecomp} in the Appendix. 

\begin{algorithm}
\caption{$\{\h{\Pi}\} \leftarrow $ LearnMixedMembership$(\h{G}, k,\tau_1, \tau_2)$}\label{algo:main}
\begin{algorithmic}[1]
\renewcommand{\algorithmicrequire}{\textbf{Input: }}
\renewcommand{\algorithmicensure}{\textbf{Output: }}
\REQUIRE Hyper-adjacency matrix $\h{G}\in \R^{|U|\cdot |T|\times |R|}$, $k$ is the number of communities,   and $\tau_1, \tau_2$ are thresholds for rank test.
\ENSURE Estimates of the community membership vectors $\Pi$.
\STATE Partition the resource set $R$ randomly into two parts  $X$, $Y$.
\STATE $\tl{R}=$Pure Resource Nodes Detection$(X, Y, U, T)$.

\STATE $\h{\Pi} \leftarrow$ TensorDecomp$(\hG(\{U, T\},\cdot), \tl{R})$

\STATE Return $\h{\Pi}$.
\end{algorithmic}
\end{algorithm}

\floatname{algorithm}{Procedure}
\begin{algorithm}
\caption{ Pure Resource Nodes Detection}\label{algo:pure}
\begin{algorithmic}[1]
\renewcommand{\algorithmicrequire}{\textbf{Input: }}
\renewcommand{\algorithmicensure}{\textbf{Output: }}
\REQUIRE $X, Y ,U, T$.
\STATE Construct Projection matrix $\Projhat= M_k M_k^\top$, where $\ksvd(\h{G}(\{U,T\}, Y))= M_k \Lambda_k V_k^\top$.
\STATE Set of pure nodes $\tl{R}\leftarrow \emptyset$.
\FOR{$x \in X$}
\IF{$\sigma_1(\mat(\Projhat\hG(\{U,T\},x)))>\tau_1$ and $ \sigma_2(\mat(\Projhat\hG(\{U,T\},x)))< \tau_2$}%
\STATE $\tl{R}\leftarrow\tl{R}\cup \{x\}$. \COMMENT{Note $\mat(\Projhat\hG(\{U,T\},x))\in \R^{|U|\times |T|}$ is   matricization}
\ENDIF
\ENDFOR
\STATE Interchange roles of $X$ and $Y$ and find pure nodes in $Y$.
\STATE Return $\tl{R}$.
\end{algorithmic}
\end{algorithm}

\section{Analysis of the Learning Algorithm} \label{sec:special}


\textbf{Notation: }Let $\tl{O}(\cdot)$ denote $O(\cdot)$ up to poly-log factors.
We use the term high probability to mean with probability $1-n^{-c}$ for any constant $c>0$.

\subsection{Assumptions}\label{sec:assume}





For simplicity, we assume that the community memberships of resources, tags and users are drawn from the same distribution. Further, we consider equal expected community sizes, i.e. $\Ebb[\pi]=1/k \cdot 1^\top$.  Additionally, we assume that the community connectivity matrices $P$, $\tl{P}$ are homogeneous\footnote{Our results can be easily extended to the case when $P$ and $\tilde{P}$ are full rank.} and equal
\beq\label{eqn:homoP} P  =\tl{P}= (p-q) I + q 11^\top.\eeq
These simplifications are merely for convenience, and can be easily removed.


\paragraph{Requirement for success of rank test: }We require that\footnote{$\tilde{\Omega}, \tilde{O}$ represent $\Omega, O$ up to poly-log factors.}
\beq \label{eqn:assume-rank} n = \tilde{\Omega}\left( \sigma_k(\Ebb[\pi\pi^\top])^{-3}\cdot \kappa(\Ebb[\pi \pi^\top])^{-2}\cdot \left(\frac{(p-q)/k+q}{(p-q)/\sqrt{k} + q}\right)^2\right) ,\eeq
where $\kappa(\cdot)$ denotes the condition number and $\sigma_k(\cdot)$ denotes the $k^{\tha}$ singular value.

We assume that $\max_{i \in [k]}\pi_x(i)=1-\epsilon, \epsilon=O(1)$ and hence there exists no node such that   their $\pi$ is between 1 and $\pi_{\max}$. 

\paragraph{Requirement for success of tensor decomposition: }


Recall that the tensor method uses only pure resource nodes. Let $\rho$ be the fraction of such pure resource nodes. Let $ w_i := \Pbb[ \pi_r(i) =1| r \in \tl{R}]$. For simplicity, we assume that $w_i\equiv1/k$. Again, this can be easily extended.

We require
the separation in edge connectivity   $p-q$ to satisfy\beq \label{eqn:sep}\frac{(p-q)^2}{p} = \tilde{\Omega} \left(   \frac{\sqrt{k} }{\sqrt{n \rho }\cdot \sigma_k(\Ebb[\pi\pi^\top])} \right).
\eeq Intuitively this implies that there should be enough separation between connectivity within a community and connectivity across communities.

\paragraph{Dependence on $p$, $q$: }Note that for the rank test,~\eqref{eqn:assume-rank}, in the well-conditioned setting we have $\sigma_k(\Ebb[\pi \pi^\top])=O(1/k)$.  Then if $p \simeq q$, we need $n= \tilde{\Omega}\left( k^3\right)$. For the case where $q < p/k$, we will require $n=\tilde{\Omega}\left( k^2\right)$. This is intuitive as  the role of $q$ is to make the components non-orthogonal, i.e., $q$ acts as noise. Therefore, smaller $q$ results in better guarantees.
For the tensor decomposition method,~\eqref{eqn:sep}, in the well-conditioned setting, if we have $n= \tilde{\Omega}\left( k^3\right)$, this means $p, q$ are constants. Alternatively, for sparse graphs, we want $p, q$ to decay. According to the constraints, we need a larger $n$. This is intuitive as in case of sparse graphs we need fewer observations and less information about unknown community memberships. Therefore, we need more samples.



Note that~\citet{AnandkumarEtal:community12} require $n=O(k^2)$ while we need $n=O(k^3)$. The reason is that we are estimating a hypergraph (they estimate a graph) and we are estimating more parameters in this model. Therefore, we need more samples.



\subsection{Guarantees}

We now establish main results on recovery at the end of our algorithm. We first show that under the assumptions in the previous section, we obtain an $\ell_2$ guarantee for recovery of the membership weights of  source nodes in each community. We should note that this result can be extended to recovery of membership for tag and user nodes as well. In this case, there will be additional perturbation terms.

Let  $\tilde{\Pi}$ be the reconstruction of communities (of resources, users and tags)  using the tensor method in Algorithm~\ref{alg:tensordecomp} in the Appendix, but  before thresholding. For a matrix $M$, let $(M)^i$ denote the $i^{\tha}$ row. Recall that $(\Pi)^i$ denotes the memberships of all the nodes in the $i^{\tha}$ community, since $\Pi\in \Rbb^{(|R|+|U|+|T|)\times k}$.
We have the following result:

\begin{theorem}[Reconstruction of communities (before thresholding)] \label{thm:one}
 We have w.h.p. \beq\epsilon_\pi:= \max_{i\in [k]}
\| (\tilde{\Pi})^i - (\Pi)^i\|_2 =\tilde{O}\left(\frac{\sqrt{k}\cdot p\cdot \kappa(\Ebb[\pi \pi^\top])}{\sqrt{\rho  }(p-q)^2} \right) .  \eeq
\end{theorem}

\paragraph{Remark: }Note that the $\ell_2$ norm  above is taken  over all the nodes of the network and we expect this to be $O(\sqrt{n})$ if error at each node is $O(1)$. Assuming $\Ebb[\pi\pi^\top]$ is well conditioned and when $\rho, p,q=\Omega(1)$, we get a better guarantee that $\epsilon_{\pi}=O(\sqrt{k})$.

Now we further show that when the distribution of $\pi$ is ``mostly'' sparse, i.e. each node's membership vector does not have too many large entries, we can improve the above $\ell_2$ guarantees into $\ell_1$ guarantees via thresholding.

Specifically, assuming that  the distribution of $\pi$ satisfies
\[\Pbb[\pi(i)\geq \tau] \leq \frac{C}{k} \log(1/\tau), \quad \forall i \in [k]\] for $\tau =O(\epsilon_{\pi}\cdot\frac{k}{n})$, we have the following result. This is equivalent to the case that the tail $\tau$ is exponentially small in $k$, i.e., sparsity.

\paragraph{Remark: }Dirichlet distribution satisfies this assumption when $\sum_i \alpha_i < 1$, where $\alpha_i$ represent the Dirichlet concentration parameters.

\begin{theorem}[$\ell_1$ guarantee for reconstruction after thresholding] \label{thm:two}
We have \beq \| \hat{\Pi^i}-\Pi^i\|_1=\tilde{O}\left(\epsilon_{\pi}\cdot \sqrt{\frac{n}{k}}\right)=\tilde{O}\left(\frac{\sqrt{n}\cdot p\cdot \kappa(\Ebb[\pi \pi^\top])}{\sqrt{\rho  }(p-q)^2} \right) ,\eeq where $\hat{\Pi^i}$ is the result of thresholding with $\tau=O(\epsilon_{\pi}\cdot\frac{k}{n})$.
\end{theorem}

\paragraph{Remark: }Note that the $\ell_1$ norm  above is taken  over all the nodes of the network and we expect this to be $O({n})$ if error at each node is $O(1)$. Assuming $\Ebb[\pi\pi^\top]$ is well conditioned and when $\rho, p,q=\Omega(1)$, we get a better guarantee of $O(\sqrt{n})$. Hence, we obtain good error guarantees in both cases on $\ell_1$ and $\ell_2$ norms.

For proof of the Theorems, see Appendix~\ref{app:proof}.

\section{Overview of Proof}

\subsection{Analysis of Graph Moments under MMSF}
\subsubsection{Overview of Kronecker and Khatri-Rao products: }We require the  notions of Kronecker $A \otimes B$ and  Khatri-Rao products $A \odot B$ between two matrices $A$ and $B$. First we define the {\em Kronecker product} $A \otimes B$ between matrices $A \in \Rbb^{n_1\times k_1}$ and $B\in \Rbb^{n_2\times k_2}$. Its $(\bfi, \bfj)^{\tha}$ entry is given by
\[ (A \otimes B)_{\bfi, \bfj} :=A_{i_1, j_1} B_{i_2,j_2}, \quad \bfi=\{i_1,i_2\}\in [n_1]\times [n_2], \bfj=\{j_1,j_2\}\in [k_1]\times [k_2].\]
Thus, for two vectors $a$ and $b$, we have
\[ (a \otimes b)_{\bfi} := a_{i_1} b_{i_2}, \quad \bfi=\{i_1,i_2\}\in [n_1]\times [n_2].\]

For the Khatri-Rao product  $A \odot B$ between matrices  $A \in \Rbb^{n_1\times k}$ and $B\in \Rbb^{n_2\times k}$, we have its
 $(\bfi, j)^{\tha}$ as
\[ A\odot B(\bfi, j) := A_{i_1, j} B_{i_2, j}, \quad \bfi=\{i_1, i_2\}\in [n_1]\times[ n_2], j \in [k].\] In other words, we have
\[ A \odot B := [ a_1\otimes b_1\quad a_2\otimes b_2 \quad\ldots a_k \otimes b_k],\] where $a_i , b_i$ are the $i^{\tha}$ columns of $A$ and $B$.
Note the difference between the Kronecker and the Khatri-Rao products. While the Kronecker product expands both the number of rows and columns, the Khatri-Rao product preserves the original number of columns. We will also use another simple fact that
\beq (A\otimes B)(C \otimes D) = AC \otimes BD\label{eqn:kronecker-result}.\eeq

\subsubsection{Result on Correctness of the Algorithm}
Recall that $P\in [0,1]^{k\times k}$ denotes the  connectivity matrix between communities of users and resources and $\tl{P}\in [0,1]^{k\times k}$  denotes the corresponding connectivity between communities of resources and tags.
Define
\begin{equation}
F:= \Pi_U^\top P, \quad   \tl{F}:= \Pi_T^\top \tl{P}.
\label{eqn:F}
\end{equation} Let $F_u = \pi_u^\top P$ be the row vector corresponding to user $u$ and similarly $\tl{F}_t$ corresponds to tag $t$. Similarly, let $F_{A}= \Pi_A^\top P$ be the sub-matrix of $F$.

We now provide a simple result on the average hyper-edge connectivity and the form of the $3$-star counts, given the community memberships.  

\begin{proposition}[Form of Graph Moments]\label{prop:moment}Under the MMSF model proposed in Section~\ref{sec:model}, we have that the generated hyper-graph $\hG\in \R^{|U|\cdot |T|\times |R|}$ satisfies \beq \label{eqn:moment}  G:=\Ebb[\hG| \Pi] = (F \odot \tl{F}) \Pi_R ,\eeq where $\odot $ denotes the Khatri-Rao product. Moreover, for a given resource $r\in R$, the column vector $\hG(r,\{U,T\})$ has   conditionally independent entries given the community membership vector $\pi_r$.
If $\tl{R}\subset R$ is the set of (exactly) pure nodes, then the $3$-star count defined in \eqref{eqn:threestardef} satisfies
\beq \Tc_{\tl{R}\rightarrow A,B,C} := 
\Ebb[\h{\Tc}_{\tl{R}\rightarrow A,B,C}|\Pi] = \sum_{i \in [k]} w_i (H_A\otimes H_B\otimes H_C),\eeq where $w_i$ is
\[ w_i := \Pbb[ \pi_r(i) =1| r \in \tl{R}],\] and  $H_A:= F_{U(A)}\odot \tl{F}_{T(A)}$, and similarly, $H_B$ and $H_C$.
\end{proposition}  The above results
follow from modeling assumptions in Section~\ref{sec:model}, and in particular, the conditional independence relationships among the different variables. For details, see Appendix~\ref{app:moments}.

In \eqref{eqn:moment}, note that a if column of $G(X;\{U,T\})$ corresponds to a pure node $x\in X$, then the matrix has  rank of one, since $\pi_x$ corresponds to a coordinate basis vector. On the other hand, for the case where columns  correspond to mixed nodes, the matrix has rank bigger than one.
Thus, the rank criterion succeeds in identifying the pure nodes in $X$ under exact moments.

\begin{lemma}[Correctness of the method under exact moments]\label{lemma:correct}
Assume $F\odot \tl{F}$ has full column rank, and $\Pi_Y$ has full row rank, where  $Y\subset R$ is used for constructing the projection matrix,   then the proposed method LearnMixedMembership in Algorithm~\ref{algo:main} correctly learns the community membership matrix $\Pi$.\end{lemma}

\bprf Using the form of the moments in Proposition~\ref{prop:moment}, we have that if $r\in \tl{R}$ is a pure node, then
$G(r;\{U,T\}) = (F\odot \tl{F}) \pi_r$ is rank one since it selects only one column of $F\odot \tl{F}$. Thus, the
 rank test in Algorithm~\ref{algo:main} succeeds in recovering the pure nodes. The correctness of tensor method follows from~\cite{AnandkumarEtal:community12}.\eprf


Since we only have sampled graph $\h{G}$ and not the exact moments, we need to carry out perturbation analysis, which is outlined below.

\subsection{Perturbation Analysis}

Recall that $\Projhat=M_k M_k^\top$ is the projection matrix corresponding to $\ksvd(\hG(\{U,T\}, Y))= M_k \Lambda_k V_k^\top$.
Define the perturbation between empirical and exact moments upon projection as
\beq   m_x:=\|\Projhat \h{G}(\{U,T\},x) -G(\{U,T\},x)\|, \quad \forall \, x \in X,\quad \errrank:= \max_x \| m_x\|.\label{eqn:mx}\eeq 
The above perturbation can be divided into two parts
\[\|m_x\| \leq   \|\Projhat (\h{G}(\{U,T\},x) -G(\{U,T\},x))\| + \| (\Projhat - \Proj)G(\{U,T\},x)\|. \] The first term is commonly referred to as {\em distance perturbation} and the second term is the {\em subspace perturbation}. We establish these perturbation bounds below.

We begin our perturbation analysis by bounding $m_x$ as defined in Eqn.~\eqref{eqn:mx}.

\begin{lemma}[Distance perturbation]\label{lemma:distance}
Under the assumptions of Section~\ref{sec:assume}, with probability $1-\delta$, we have for all $x\in X$,
 \[\|\Projhat (\h{G}(\{U,T\},x) -G(\{U,T\},x))\|\leq \sqrt{k} p  \left(1 +\frac{C'}{\sqrt{k}}\left(\log(n/\delta)\right)^4 \right)^{1/2} ,\] for some constant $C'>0$.
\end{lemma}


See Appendix~\ref{proof:distance} and Appendix~\ref{proof:subspace} for details. Notice that the subspace perturbation dominates.

\begin{lemma}[Subspace perturbation]\label{lemma:subspace}
We have the subspace perturbation as\[\| (\Projhat- \Proj)G(\{U,T\},x)\|\leq 2 \sigma^{-1}_k(\Pi_Y)\sqrt{\|F\odot \tl{F}\|_1}. \] Under the assumptions of Section~\ref{sec:assume}, w.h.p. this reduces as  \[\| (\Projhat- \Proj)G(\{U,T\},x)\|\leq O\left(\frac{\sqrt{n}}{\sqrt{\sigma_k(\Ebb[\pi \pi^\top])}}\cdot \left(\frac{p-q}{k}+q\right) \right).\]
\end{lemma}

See Appendix~\ref{proof:subspace}.

%

\subsection{Analysis of Rank Test}

Recall that from the perturbation analysis, we have bound $\errrank$ on the error vector $m_x$, defined in \eqref{eqn:mx}. We assume there exist no node such that $\max_{i \in [k]}\pi_x(i)$ is between the threshold given in~\eqref{eqn:pimax} and 1. 
We have the following result on the rank test.
 
\begin{lemma}[Conditions for Success of Rank Test]\label{lemma:ranktest}When the thresholds in Algorithm~\ref{algo:main} are chosen\[0< \tau_1 <\min_i \|(F_{U})_i\|\cdot \|(\tl{F}_{T})_i\|-\errrank,\quad\tau_2 > \errrank,\]
then all the pure nodes pass the rank test. Moreover, any node $x\in X$ passing the rank test satisfies
\beq \label{eqn:pimax}\max_{i \in [k]}\pi_x(i) \geq \frac{\tau_1-\tau_2-2\errrank}{\max_i \|(F_{U})_i\|\cdot \|(\tl{F}_{T})_i\|}.\eeq
\end{lemma}

\bprf See Appendix~\ref{proof:ranktest}.\eprf

The above result states that we can correctly detect pure nodes using the rank test. The conditions stem from the fact that we require the top eigen-value to pass the test and the second top eigen-value to not pass the test. For a pure node, $\sigma_1(\mat(\Projhat G(\{U_1,T_1\},x)))$ is $\min_i \|(F_{U_1})_i\|\cdot \|(\tl{F}_{T_1})_i\|$. To account for empirical error, we consider $\errrank$. In addition, the second-top eigen-value can be as small as $0$. We also note the error in empirical estimation.
This result allows us to control the perturbation in the $3$-star tensor constructed using the nodes which passed the rank test.
 


\section{Conclusion}
In this paper, we propose a novel probabilistic approach for modeling folksonomies, and  propose a guaranteed approach for detecting overlapping communities in them. We present a more scalable approach where  realistic conditional independence constraints are imposed. These constraints are natural for social tagging systems, and they lead to scalable modeling and tractable learning.
While the original MMSB model assumes that the communities are drawn from a Dirichlet distribution, here, we do not require such a strong parametric assumption.   Note that the Dirichlet assumption for community memberships can be limiting and cannot model general correlations in memberships. Here, we impose a weak assumption that   a certain fraction of resource nodes are ``pure'' and belong to a single community. This is reasonable to expect in practice. We establish that the communities are identifiable under these natural assumptions, and can be learnt efficiently using spectral approaches.
Considering future directions, we note that  social tagging assumes a specific structure. Therefore, it is of interest to extend this model to more general hypergraphs. 


\subsection*{Acknowledgment}
A.Anandkumar is supported in part by Microsoft Faculty Fellowship,  NSF Career award CCF-1254106, NSF Award CCF-1219234,  and ARO YIP Award W911NF-13-1-0084. H. Sedghi is supported by ONR Award N$00014-14-1-0665$.

The authors thank Majid Janzamin for detailed discussion on rank test analysis.
 The authors thank Rong Ge and Yash Deshpande for extensive initial discussions  during the visit of AA to Microsoft Research New England in Summer 2013 regarding the pairwise mixed membership models without the Dirichlet assumption. The authors also acknowledge detailed discussions with Kamalika Chaudhuri regarding analysis of spectral clustering.

\renewcommand{\appendixpagename}{Appendix}

\appendixpage
\appendix

\section{Moments under MMSF model and Algorithm Correctness}\label{app:moments}
\bprfof{Proposition~\ref{prop:moment}} We have
 \begin{align}\nn \Ebb[\h{G}(\{u,t\},r)|\pi_r, \pi_t, \pi_u] &\eqa  \Ebb[\Ebb[\h{G}(\{u,t\},r)|z_{r \rightarrow \{u,t\}}, \pi_t,\pi_r, \pi_u]]\\ \nn&\eqb \Ebb[\Ebb[\h{B}_{r\rightarrow u;t} \cdot \h{B}_{r\rightarrow t;u}|z_{r \rightarrow \{u,t\}},\pi_t, \pi_u]|\pi_r]\\
&\eqc\Ebb[F_u z_{r \rightarrow \{u,t\} }\cdot \tl{F}_tz_{r \rightarrow \{u,t\}} |\pi_r],\label{eqn:int-result-moments}\end{align}
where $(a)$ and $(b)$ are from the  assumption \eqref{eqn:hyper} that \[ \h{G}(\{u,t\},r)=\h{B}_{r\rightarrow u;t} \cdot \h{B}_{r\rightarrow t;u} , \] where $\h{B}_{r\rightarrow u;t}$ and $\h{B}_{r\rightarrow t;u}$ are Bernoulli draws, which only depend on the contextual variables $z_{r \rightarrow \{u,t\}}, z_{u\rightarrow \{r,t\}}$ and $z_{t \rightarrow \{u,r\}}$, and therefore $\hG(\{u,t\},r)-z_{r \rightarrow \{u,t\}}-\pi_r$ form a Markov chain. This also establishes that   $\h{G}(\{u,t\},r)$ and $\h{G}(\{u',t'\},r)$ are    conditionally independent given the community membership vector $\pi_r$, for $u \neq u'$ and $t \neq t'$.

For (c), we have that
\begin{align*} \Ebb[\h{B}_{r\rightarrow u;t} |z_{r \rightarrow \{u,t\}}, \pi_u] &= \Ebb[\Ebb[\h{B}_{r\rightarrow u;t} |z_{r \rightarrow \{u,t\}}, z_{u \rightarrow \{r,t\}}]| \pi_u]\\ &=\Ebb[ z_{u \rightarrow \{t,r\}}^\top  P  z_{r \rightarrow \{u,t\}}| z_{r \rightarrow \{u,t\}},\pi_u]\\ &= \pi_u^\top P z_{r \rightarrow \{u,t\}} \\ &= F_u z_{r \rightarrow \{u,t\}}
\end{align*} from  \eqref{eqn:bern} and the fact that
\[ \Ebb[z_{u \rightarrow \{t,r\}} |\pi_u] = \pi_u.\]

Thus, we have
\begin{align*} \Ebb[\h{G}(\{U,T\},r)|\pi_r, \Pi_T, \Pi_U] &\eqa \Ebb[F z_{r \rightarrow \{u,t\}}\otimes \tl{F}z_{r \rightarrow \{u,t\}} |\pi_r]\\&\eqb\Ebb[ ( F\otimes \tl{F})( z_{r \rightarrow \{u,t\} }\otimes z_{r \rightarrow \{u,t\} }) |\pi_r]\\ &\eqc  \sum_{i \in [k]} \pi_r(i)( F\otimes \tl{F}) (e_i \otimes e_i)\\ &\eqd (F\odot \tl{F}) \pi_r, \end{align*} where (a) follows from \eqref{eqn:int-result-moments} and
(b) follows from the fact \eqref{eqn:kronecker-result}. (c) follows from the fact that $z_{r \rightarrow \{u,t\} }$ takes value $e_i$ with probability $\pi_r(i)$, where $e_i\in \R^k$ is the basis vector in the $i^{\tha}$ coordinate. (d) follows from the definition of Khatri-Rao product.

The form of the $3$-star moment is from the lines of~\cite[Prop 2.1]{AnandkumarEtal:community12}, and relies on the assumption that $\tl{R}$ consists of pure nodes.

\eprfof\\

\section{ Learning using Tensor Decomposition} \label{app:tensor}
We now recap the tensor decomposition approach proposed in~\cite{AnandkumarEtal:community12} here. This is shown in Algorithm~\ref{alg:tensordecomp} with modifications specific to our framework.


We partition $U, T$ into three sets for the different tasks explained in the Algorithm~\ref{alg:tensordecomp}. Also note that with knowledge of community connectivity matrices, we can learn community memberships for mixed resource nodes as well. 

\begin{algorithm}[h]
\caption{$(\h{\Pi}) \leftarrow$ TensorDecomp$(\hG, \tl{R})$}\label{alg:tensordecomp}
\begin{algorithmic}
\renewcommand{\algorithmicrequire}{\textbf{Input: }}
\renewcommand{\algorithmicensure}{\textbf{Output: }}
\STATE Let $P\in \Rbb^{k \times k}$ be the community connectivity matrix from user communities to resource communities and similarly $\tilde{P}$ is connectivity from tag communities to resource communities. $\tilde{R}$ are  estimated pure resource nodes. Partition $\{U,T\}$ into $\{U_i, T_i\}$ for $i=1, 2,3$.
\STATE Compute whitened and symmetrized tensor $\Tc\leftarrow \hat{G}_{\tilde{R}\rightarrow \{A,B,C\}}(\h{W}_A, \h{W}_B \h{S}_{AB},\h{W}_C \h{S}_{AC})$, where $A,B, C$ form a partition of $\{U_2, T_2\}$. Use $\{U_3, T_3\}$ for computing the whitening matrices.
\STATE $\{\h{\lambda}, \h{\Phi}\}\leftarrow $TensorEigen$(T, \{\h{W}^\top_A \h{G}^\top_{i, A}\}_{i \notin A}, N)$.
\COMMENT{$\h{\Phi}$ is a $k\times k$ matrix with each columns being an estimated eigenvector and $\h{\lambda}$ is the vector of estimated eigenvalues.}
\STATE     $\h{\Pi}_{R}  \leftarrow\thres( \Diag(\h{\lambda})^{-1}\h{\Phi}^\top \h{W}_A^\top \hG_{R,A}^\top\,,\,\, \tau)$.
\RETURN $(\h{\Pi})$.
\end{algorithmic}
\end{algorithm}

\begin{algorithm}[h]
\caption{$\{\lambda, \Phi\}\leftarrow $TensorEigen$(T,\, \{v_i\}_{i\in [L]}, N)$~\cite{AnandkumarEtal:community12}}\label{alg:robustpower}
\begin{algorithmic}
\renewcommand{\algorithmicrequire}{\textbf{Input: }}
\renewcommand{\algorithmicensure}{\textbf{Output: }}
\REQUIRE Tensor $T\in \R^{k \times k \times k}$, $L$ initialization vectors $\{v_i\}_{i\in L}$, number of
iterations  $N$.
\ENSURE the estimated eigenvalue/eigenvector pairs $\{\lambda, \Phi\}$, where $\lambda$ is the vector of eigenvalues and $\Phi$ is the matrix of eigenvectors.

\FOR{$i =1$ to $k$}
\FOR{$\tau = 1$ to $L$}
\STATE $\theta_0\leftarrow v_\tau$.
\FOR{$t = 1$ to $N$}
\STATE $\tilde{T}\leftarrow T$.
\FOR{$j=1$ to $i-1$ (when $i>1$)}
\IF{$|\lambda_j \inner{\theta_{t}^{(\tau)}, \phi_j}|>\xi$}
\STATE $\tilde{T}\leftarrow \tilde{T}- \lambda_j \phi_j^{\otimes 3}$.
\ENDIF
\ENDFOR

\STATE Compute power iteration update
$
\theta_{t}^{(\tau)}  :=
\frac{\tilde{T}(I, \theta_{t-1}^{(\tau)}, \theta_{t-1}^{(\tau)})}
{\|\tilde{T}(I, \theta_{t-1}^{(\tau)}, \theta_{t-1}^{(\tau)})\|}
$\ENDFOR
\ENDFOR

\STATE Let $\tau^* := \arg\max_{\tau \in L} \{ \tilde{T}(\theta_{N}^{(\tau)},
\theta_{N}^{(\tau)}, \theta_{N}^{(\tau)}) \}$.

\STATE Do $N$ power iteration updates starting from
$\theta_{N}^{(\tau^*)}$ to obtain eigenvector estimate $\phi_i$, and set $\lambda_i :=
\tilde{T}(\phi_i, \phi_i, \phi_i)$.

\ENDFOR

\RETURN the estimated eigenvalue/eigenvectors
$(\lambda, \Phi)$.

\end{algorithmic}
\end{algorithm}

%

\section{Perturbation Analysis: Proof of Theorems~\ref{thm:one},~\ref{thm:two}} \label{app:proof}

\paragraph{Notation: }For a vector $v$, let $\|v\|$ denote its $2$-norm. Let $\Diag(v)$ denote a diagonal matrix with diagonal entries given by a vector $v$. For a matrix $M$, let $(M)_i$ and $(M)^i$ denote its $i^{\tha}$ column and row respectively. Let $\|M\|_1$ denote column absolute sum and $\|M \|_\infty$ denote row absolute sum of $M$.  Let $M^\dagger$ denote the Moore–Penrose pseudo-inverse of $M$.

\subsection{Distance Concentration: Proof of Lemma~\ref{lemma:distance}}\label{proof:distance}

The proof is along the lines of~\cite[Theorem 13]{McSherry01} but we apply   Hanson-Wright bound in Proposition~\ref{prop:hansonwright} to get a better perturbation guarantee without the need for constructing the so-called combinatorial projection, as in~\cite{McSherry01}.


We have $h_x:=\hG(x ; \{U,T\}) -G(x ;\{U,T\})$ and let $\sigma^2 = \max_{i} \Ebb[h_x(i)^2|\pi_x]$.
Note the simple fact
\[ \|\Projhat h_x\|^2 =  h_x^\top  \Projhat^2 h_x  =  h_x^\top  \Projhat h_x ,\] since $\Projhat$ is a projection matrix. From Proposition~\ref{prop:moment}, we have that the entries of $h_x$ are conditionally independent given $\pi_x$. Thus, the Hanson-Wright inequality  in Proposition~\ref{prop:hansonwright} is applicable, and we have
with probability $1-\delta$, for all $x\in X$,
\beq\label{eqn:meanvariance} h_x^\top  \Projhat h_x  \leq \Ebb[h_x^\top \Projhat h_x|\pi_x]
+ C'\sigma^2 \|\Projhat\|_{\Fbb}\left(\log(n/\delta) \right)^4   \eeq
Now $\|\Projhat\|_{\Fbb} \leq \sqrt{k} \|\Projhat\|=\sqrt{k}$.
The expectation is
\[  \Ebb[h_x^\top \Projhat h_x|\pi_x] \leq \Tr(\Projhat) \sigma^2= k \sigma^2, \] using the property that $\Projhat$ is idempotent. Thus, we have from \eqref{eqn:meanvariance}, with probability $1-\delta$, for all $x\in X$,
\[ h_x^\top  \Projhat h_x  \leq  k \sigma^2
+ C'\sqrt{k}\sigma^2  \left(\log(n/\delta) \right)^4  , \] and we see that the mean term dominates and the bound is $\tl{O}(k \sigma^2)$.

Draw random variables
 \begin{align*}
  \nn\h{B}_{r\rightarrow u;t} &\sim \mbox{ Bernoulli}(  z_{u \rightarrow \{t,r\}}^\top  P  z_{r \rightarrow \{u,t\}})\\ \h{B}_{r\rightarrow t;u} &\sim\mbox{ Bernoulli}(  z_{t \rightarrow \{u,r\}}^\top  \tilde{P} z_{r \rightarrow \{u,t\}}).\end{align*}
  The presence of  hyper-edge $G(\{u,t\},r)$ is given by the product
 \begin{align*} \h{G}(\{u,t\},r) =\h{B}_{r\rightarrow u;t} \cdot \h{B}_{r\rightarrow t;u}.\end{align*}

The variance    is on lines of proof of Lemma~\ref{lemma:2starcountconc} and we repeat it here.
\begin{align*}\max_{i}\Ebb[h_x(i)^2|\pi_x]&=\max_{u\in U, v \in V}\Ebb[\hB_{x\rightarrow u;t}\hB_{x\rightarrow t;u}-((F\odot \tl{F})\pi_x)_{ut}]^2\\& \leq\max_{u\in U, v \in V} ((F\odot \tl{F})\pi_x)_{ut},\\ &\leq \max_{u\in U, v\in V} \sum_{j\in [k]} F(u,j) \tl{F}(t,j) \pi_x(j) \\ \nn&
\leq   \max_{i\in [k]} \sum_{ j\in [k]} P(i,j)\tl{P}(i,j) \pi_x(j) \\
&\leq  P_{\max}^2\end{align*}

\subsection{Proof of Lemma~\ref{lemma:subspace}}\label{proof:subspace}
From Davis-Kahan in Proposition~\ref{prop:daviskahan}, we have
\[\| (\Projhat- I)G(\{U,T\},Y)\| \leq 2 \| \hG(\{U,T\},Y)-G(\{U,T\},Y)\|. \] and thus
\[ \| (\Projhat- I)G(\{U,T\},x)\|\leq 2 \| \hG(\{U,T\},Y)-G(\{U,T\},Y)\|\cdot \| G(\{U,T\},Y)^\dagger \cdot G(\{U,T\},x)\|\]
Now, \[ G(\{U,T\},Y)^\dagger  =\left( (F \odot \tl{F}) \Pi_Y\right)^\dagger = \Pi_Y^\dagger (F \odot \tl{F})^\dagger ,\]
since the assumption is that $F \odot \tl{F}$ has full column rank and $\Pi_Y$ has full row rank. Thus, we have
\[ G(\{U,T\},Y)^\dagger \cdot G(\{U,T\},x) = \Pi_Y^\dagger (F \odot \tl{F})^\dagger (F \odot \tl{F}) \pi_x= \Pi_Y^\dagger  \cdot \pi_x,\] since    $(F \odot \tl{F})^\dagger (F \odot \tl{F}) = I$ due to full column rank, when $|U|$ and $|T|$ are sufficiently large, due to concentration result from Lemma~\ref{lemma:spectral}. Note that under assumption A3, the variance terms in Lemma~\ref{lemma:spectral} are decaying and we have that  $F \odot \tl{F}$ has full column rank w.h.p.
From Lemma~\ref{lemma:2starcountconc}, we have the result.


\subsection{Analysis of Rank Test: Lemma~\ref{lemma:ranktest}}\label{proof:ranktest}

Consider the test  under expected moments $G:=\Ebb[\h{G}|\Pi]$.  For every node $x\in X$ ($R$ is randomly partitioned into $X, Y$), which passes the rank test in Algorithm~\ref{algo:main}, by definition,
\[ \|\mat( \hat{G}(\{U,T\},x))\|>\tau_1, \quad\mbox{and}\quad  \sigma_2(\mat( \hat{G}(\{U,T\},x)))< \tau_2.\]


We use the following approximation.
\begin{align*}
\Vert F_i \Vert \simeq \sqrt{(p-q)^2 \Vert \Pi^i\Vert^2+nq^2+2(p-q)q \Vert \Pi^i\Vert_1}
\end{align*}

Recall the form of $G$ from Proposition~\ref{prop:moment}
\[ \mat( G(\{U,T\},x))= F_{U} \Diag(\pi_x) \tl{F}_{T}^\top.\]

First we consider the case, $p \simeq q$. Following lines of~\citet{anandkumar2014guaranteed}, we have that
\[ \vert \sigma_1 - \pi_{\max}n(\frac{p-q}{k}+q)^2 \vert \leq \Vert E \Vert + \errrank \] where
\[ \Vert E \Vert \leq \sqrt{k} \pi_{2,\max}n(\frac{p-q}{k}+q)^2 \frac{\Vert \Ebb[\pi \pi^\top]\Vert q^2}{\left( \frac{p-q}{k}+q \right)^3} \left\lbrace p \sqrt{\Ebb[\pi \pi^\top]}+\frac{\Vert \Ebb[\pi \pi^\top]\Vert q^2}{\left( \frac{p-q}{k}+q \right)} \right\rbrace. \]
Hence, we have that
\[ \label{eqn:sigma2} \sigma_2 \geq \pi_{2,\max}n(\frac{p-q}{k}+q)^2 - \Vert E \Vert- \errrank -(1/\tilde{\mu}) \errrank-\pi_{3,\max} np^2 \Vert \Ebb[\pi \pi^\top]\Vert,
\]
where we assume $\pi_{\max} \geq (1+\mu)\pi_{2,\max}$ and $\tilde{\mu}:=\frac{1+\mu-\mu_R-\mu_E}{1+\mu}$, $\mu_R :=\frac{\Vert F \Vert}{\Vert F_i \Vert}$
, $\mu_E :=\frac{\Vert E \Vert}{\pi_{2,\max}n(\frac{p-q}{k}+q)^2}$.

We note that $\errrank$ dominates $\Vert E \Vert$ and the last term. Therefore,
\[
\tau_2-\errrank \geq  \sigma_2(\mat( G(\{U,T\},x)))\geq \pi_{2,\max}n(\frac{p-q}{k}+q)^2-(1+1/\tilde{\mu})\errrank
,\]
and
\begin{align*} \tau_1+\errrank &\leq \|F_{U} \Diag(\pi_x) \tl{F}_{T}^\top\|\\ &\leq \pi_{\max} \max_i \|(F_{U_1})_i\|\cdot \|(\tl{F}_{T})_i\| + \pi_{2,\max}n(\frac{p-q}{k}+q)^2 \\ &\leq \pi_{\max} \max_i \|(F_{U})_i\|\cdot \|(\tl{F}_{T})_i\| + \tau_2+1/\tilde{\mu}\errrank. \end{align*}

Combining we have that any vector which passes the rank test satisfies \[ \pi_{\max} \geq \frac{\tau_1-\tau_2+(1-1/\tilde{\mu})\errrank}{\max_i \|(F_{U})_i\|\cdot \|(\tl{F}_{T})_i\|}.\]

Now, for the case where $q < p/k$, the bound on $\Vert E \Vert$ is almost 0, $\mu_R \simeq 1$ and $\mu_E =0$. Hence Eqn.~\eqref{eqn:sigma2} always holds. This is intuitive as  the role of $q$ is to make the components non-orthogonal, i.e., $q$ acts as noise. Therefore, smaller $q$ results in better guarantees.

With $|U| = |T|=\Theta(n)$, and  using the concentration bounds in Lemma~\ref{lemma:spectral}, we have that with probability $1-\delta$, \[ \| (F_{U})_i\| \cdot \| (\tl{F}_{T})_i\| =O\left( \sqrt{|U|\cdot |T|} \|\Ebb[\pi\pi^\top]\|\cdot(p-q +\sqrt{k}q)\right)\] assuming homegenous setting.



For $\errrank$, the subspace perturbation dominates. From Lemma~\ref{lemma:spectral}, we have
\[ \| F \odot \tl{F}\|_1 = O\left( n^2
\left(\frac{p-q}{k}+q\right)^2\right).\] Thus, we have
the subspace perturbation from Lemma~\ref{lemma:subspace} as
\[\errrank = O\left(\frac{\sqrt{n} }{\sqrt{\sigma_k(\Ebb[\pi\pi^\top])}}\cdot \left(\frac{p-q}{k}+q \right) \right). \]Substituting for the condition that $\tau_1 = \Omega(\errrank)$, we obtain assumption \eqref{eqn:assume-rank}. Thus, the rank test succeeds in this setting.
\subsection{Perturbation Analysis for the Tensor Method}

This is along the lines of analysis in~\cite{AnandkumarEtal:community12}. However, notice here due to hypergraph setting, we need to redo the individual perturbations.
Recall that $w_i:= \Pbb[i=\argmax_j \pi(j)| \pi\mbox{ is pure}]$ and $\rho=\Pbb[  \pi\mbox{ is  pure}]$. The size of recovered set of pure nodes $\tl{R}= \Theta(n \rho)$, assuming $n \rho>1$.

We provide the perturbation of the whitened tensor. Let $\Phi:=W_A^\top H_A \Diag(\eta)^{1/2}$ be the eigenvectors of the whitened tensor under exact moments and $\lambda:= \Diag(\eta)^{-1/2}$ be the eigenvalues. $S, \h{S}$ respectively denote the exact and empirical symmetrization matrix for different cases based on their subscript.

\begin{lemma}[Perturbation of whitened tensor]We have w.h.p.
\begin{align} \epsilon_{\Tc}:=& \left\|\h{\Tc}_{\tl{R}\rightarrow \{A,B,C\}}(\hat{W}_A, \hat{W}_B \h{S}_{AB}, \hat{W}_C \h{S}_{AC})- \sum_{i \in [k]} \lambda_i\Phi^{\otimes 3}\right\|\nn \\ =& O\left(\frac{p}{\sqrt{n\rho  }w_{\min}\cdot(p-q)^2\cdot \sigma_k(\Ebb[\pi \pi^\top])} \right)\end{align}
\end{lemma}

\bprf Let $\Tc:= \Ebb[\h{\Tc}| \Pi_{A,B,C}]$.
\begin{align*}\epsilon_1&:= \left\|\h{\Tc}(\hat{W}_A, \hat{W}_B \h{S}_{AB}, \hat{W}_C \h{S}_{AC}) - \Tc(\hat{W}_A, \hat{W}_B \h{S}_{AB}, \hat{W}_C \h{S}_{AC})\right\|\\
\epsilon_2&:=\left\|\Tc(\hat{W}_A, \hat{W}_B \h{S}_{AB}, \hat{W}_C \h{S}_{AC})- \Tc(W_A, W_B S_{AB}, W_C  S_{AC})\right\|
\end{align*}

For $\epsilon_1$, the dominant term in the perturbation bound is
\begin{align*}& O \left( \frac{1}{|\tl{R}|}\|\tl{W}^\top_B H_B\|^2\left\|
\sum_{i\in Y}\left(\h{W}^\top_A   (\h{G}_{A,i} -   H_A \pi_i)\right)\right\| \right)\\&=  O \left( \frac{1}{ w_{\min}}
\frac{1}{|\tl{R}|} \left\|
\sum_{i\in Y}\left(\h{W}^\top_A   (\h{G}_{A,i} -   H_A \pi_i)\right)\right\| \right)  \end{align*}


The second term is
\[\epsilon_2\leq \frac{\epsilon_W}{\sqrt{w_{\min}}},\] since due to whitening property.


Now imposing the requirement that \[ \epsilon_i < \Theta\left( \lambda_{\min} r^2 \right),\] from Theorem 11~\citep{AnandkumarEtal:community12}, $\lambda_{\min} = 1/\sqrt{w_{\max}}$, and we have $r=\Theta(1)$ by initialization using whitened neighborhood vectors (from lemma 25~\citep{AnandkumarEtal:community12}). $\epsilon_1$ is not the dominant error, on lines of~\citep{AnandkumarEtal:community12}. Now for $\epsilon_2$, we require \[ \epsilon_W \leq \sqrt{\frac{w_{\min}}{w_{\max}} } \leq 1,\] and using Lemma~\ref{lemma:whiten}, we have
\[ \frac{(p-q)^2}{p} \geq \frac{\sqrt{w_{\max}}}{w_{\min}} \cdot \frac{1}{\sqrt{n \rho}\cdot \sigma_k(\Ebb[\pi\pi^\top])}.\]

\eprf

\begin{lemma}[Whitening Perturbation]\label{lemma:whiten}
We have the perturbation of the whitening matrix $\hW_A$ as w.h.p.
\[ \epsilon_W:= \| \Diag(\vec{w})^{1/2} H_A^\top (\hW_A-W_A)\|= O\left(\frac{p}{\sqrt{n\rho w_{\min} }\cdot(p-q)^2\cdot \sigma_k(\Ebb[\pi \pi^\top])} \right).\]
\end{lemma}

\bprf
From~\cite[Lemma 17]{AnandkumarEtal:community12}, the whitening perturbation under the tensor method is given by
\[ \epsilon_{W}:= \| \Diag(w)^{1/2} H_A^\top (\hW_A-W_A)\|
= O\left(\frac{\epsilon_G}{\sigma_{\min}(G_{\tl{R},A})} \right) .\]

Using the bounds from Section~\ref{app:graphconc}, we have
\[  \epsilon_G:=\| \hG(\{U,T\},\tl{R})-G(\{U,T\},\tl{R})\|=O(\sqrt{\|F\odot \tl{F}\|_1})= O\left(n \left(\frac{p-q}{k}+q\right)\right),\]
and \begin{align*} \sigma_{\min}(G_{\tl{R},A}) &=\Omega\left(\sqrt{|\tl{R}| w_{\min}}\cdot\sigma_{\min}(H_A)  \right) \\ &=\Omega\left(\sqrt{ n\cdot \rho w_{\min} }\cdot \sigma_{\min}(H_A) \right).\end{align*}
From Lemma~\ref{lemma:KR}, we have \[ \sigma_{\min}(H_A)= \sigma_{\min}(F_A\odot \tl{F}_A)  = \Omega\left(n (p-q)^2\min_{i, j \neq i}\left(\Ebb[\pi_i^2] -\Ebb[\pi_i \pi_j]\right)\right).
\]Finally note that $\sigma_k(\Ebb[\pi \pi^\top])= \Theta\left( \min_{i, j \neq i}\left(\Ebb[\pi_i^2] -\Ebb[\pi_i \pi_j]\right)\right)$. Substituting we have the result. \eprf


Let $\tilde{\Pi}_Z$ be the reconstruction after the tensor method (before thresholding) on resource subset $Z \subset R - \tilde{R}$ (we do not incorporate $\tilde{R}$ to avoid dependency issues), i.e. \[ \tilde{\Pi}_Z := \Diag(\lambda)^{-1} \Phi^\top \hat{W}_A^\top G^\top_{Z,A}.\]

\begin{lemma}[Reconstruction of communities (before thresholding)]
 We have w.h.p. \beq\epsilon_\pi:= \max_{i\in Z}
\| (\tilde{\Pi}_Z)^i - (\Pi_Z)^i\| = \frac{\epsilon_{\Tc}}{\sqrt{k}}\|\Pi_Z\| =O\left(\frac{\epsilon_{\Tc}}{\sqrt{k}}\cdot \sqrt{n} \|\Ebb[\pi \pi^\top]\| \right).  \eeq
\end{lemma}

\bprf This is on lines of~\cite[Lemma 13]{AnandkumarEtal:community12}. \eprf

\subsection{Concentration of Graph Moments}\label{app:graphconc}

\begin{lemma}[Concentration of hyper-edges]\label{lemma:2starcountconc}With probability $1-\delta$, given community membership vectors $\Pi$,\[ \epsilon_G:=\| \hG(\{U,T\},Y)-G(\{U,T\},Y)\|=O(\max(\sqrt{\|F\odot \tl{F}\|_1},\sqrt{  \|(P*\tl{P})\Pi_Y\|_\infty}))\]\end{lemma}

\paragraph{Remark: }When number of nodes $n$ is large enough, the first term, \viz $\sqrt{\|F\odot \tl{F}\|_1}$ dominates.\\

\bprf The proof is on the lines of~\cite[Lemma 22]{AnandkumarEtal:community12COLT} but adapted to the setting of hyper-adjacency  rather than adjacency matrices. Let $m_y:= \hG(\{U,T\},y)-G(\{U,T\},y)$ and  $M_y:= m_y e_y^\top$ and thus  \[  \hG(\{U,T\},Y)-G(\{U,T\},Y)=\sum_y M_y,\]
Note that the random matrices   $M_y$ are conditionally independent for $y\in Y$ since $m_y$ are conditionally independent given $\pi_y$, and in each vector $m_y$, the entries are independent as well.    We apply  matrix Bernstein's inequality. We have $\Ebb[M_y|\Pi]= 0$.
We compute the variances $\sum_{y\in Y} \Ebb[M_y M_y^\top|\Pi]$ and $\sum_y \Ebb[M^\top_y M_y|\Pi]$. We have that
$\sum_y  \Ebb[M_y M_y^\top|\Pi]$ only the diagonal terms are non-zero due to independence, and \beq\label{eqn:entrybound}\Ebb[M_y M_y^\top| \Pi]\leq \Diag( (F\odot \tl{F}) \pi_y)\eeq entry-wise, assuming Bernoulli random variables. Thus,
\begin{align}\nn\|\sum_{y\in Y}\Ebb[ M_y M_y^\top|\Pi]\|
&\leq \max_{u\in U, t\in T} \sum_{y\in Y, j\in [k]} F(u,j) \tl{F}(t,j) \pi_y(j) \\ \nn&= \max_{u\in U, t\in T} \sum_{y\in Y, j\in [k]} F(u,j) \tl{F}(t,j) \Pi_Y(j,y) \\ \nn
&\leq \cdot \max_{i\in [k]} \sum_{y\in Y, j\in [k]} P(i,j)\tl{P}(i,j)  \Pi_Y(j,y)
 \\ &= \|(P*\tl{P})\Pi_Y\|_\infty,\label{eqn:l1normbound}\end{align}
  where $*$ indicates Hadamard or entry-wise product.
Similarly $\sum_{y\in Y}\Ebb[M^\top_y M_y]= \sum_{y\in Y}\Diag(\Ebb[m_y^\top m_y])\leq \|(P*\tl{P})\Pi_Y\|_\infty$.  From Lemma~\ref{lemma:spectral}, we have a bound  $\|(P*\tl{P}) \Pi_Y\|_\infty$.

We now bound $\|M_y\|= \|m_y\|$ through vector Bernstein's inequality. We have for Bernoulli $\hG$,
\[\max_{u \in U, t\in T} |\hG(\{u,t\},y)-G(\{u,t\},y)|\leq 2\] and \[\sum_{u \in U, t\in T}\Ebb[\hG(\{u,t\},y)-G(\{u,t\},y)]^2 \leq \sum_{u \in U, t\in T}((F\odot \tl{F})\pi_y)_{ut}\leq \|F\odot \tl{F}\|_1.\]

Thus with probability $1-\delta$, we have
\[\|M_y\| \leq (1+\sqrt{8\log(1/\delta)})\sqrt{\|F\odot \tl{F}\|_1 \cdot} + 8/3\log(1/\delta).\]
Thus, we have the bound that $\|\sum_y M_y\| = O(\max(\sqrt{\|F\odot \tl{F}\|_1},\sqrt{  \|(P*\tl{P})\Pi_Y\|_\infty}))$. 
\eprf\\




For a given $\delta \in (0,1)$,
we assume that the sets $U, T$ and $Y\subset R$ are large enough to satisfy \begin{align*}  \sqrt{|U|\cdot |T|}&\geq \frac{8}{3}\log \frac{|U|\cdot |T|}{\delta}\\ \sqrt{|Y|}&\geq \frac{8}{3} \log \frac{|Y|}{\delta}.\end{align*}

\begin{lemma}[Concentration bounds]\label{lemma:spectral}With probability $1-\delta$,
\begin{align*}
\| F\odot \tl{F}\|_1&\leq |U|\cdot |T| \max_i (P\cdot \Ebb[\pi])_i \max_i (\tl{P}\cdot \Ebb[\pi])_i+ \sqrt{ \frac{8}{3}  |U|\cdot |T|\cdot P_{\max}^4 \cdot \log \frac{|U|\cdot |T|}{\delta}}, \\
|\| (F\odot \tl{F})_i\|&\leq \max_i \|\Pi_U\|\cdot \|\Pi_T\|\cdot  \|P_i\|\cdot \|\tilde{P}_i\|\\ &=O\left( \sqrt{|U|\cdot |T|} \|\Ebb[\pi\pi^\top]\|\cdot(p-q +\sqrt{k}q)\right),
\end{align*}for the homogeneous setting.
Similarly for subset $Y\subset R$, we have
\begin{align*}
\|\Pi_Y\Pi_Y^\top\| &\leq |Y| \cdot \| \Ebb[\pi \pi^\top] \| + \sqrt{\frac{8}{3}|Y| \cdot\|\Ebb[\pi\pi^\top]\|^2\cdot\log \frac{|Y|}{\delta} }
\\ \sigma_k(\Pi_Y \Pi_Y^\top)&\geq |Y|\cdot \sigma_k( \Ebb[\pi \pi^\top])  - \sqrt{\frac{8}{3}|Y| \cdot \|\Ebb[\pi\pi^\top]\|^2\cdot\log \frac{|Y|}{\delta} }\\ \| (P*\tl{P})^\top \Pi_Y\|_\infty&\leq |Y| \max_i (\Ebb[\pi]^\top\cdot (P*\tl{P}))_i + \sqrt{ \frac{8 }{3} |Y|\cdot P_{\max}^4\cdot \log \frac{|Y|}{\delta}}
\end{align*}\end{lemma}


\paragraph{Remark: }Note that $\sigma(P)=\Theta(p-q)$ and $\|P\|= \Theta(p+q)$ for homogeneous $P$. Under  Assumption A3, the variance terms are small and the above quantities are close to their expectation.\\

\bprf
To   bound on $\|F\odot \tl{F}\|_1$, we note that $\|\Ebb[F\odot \tl{F}] \|_1 \leq |U| \cdot |T|\max_i (P^\top\cdot  \Ebb[\pi])_i(\tl{P}^\top\cdot  \Ebb[\pi])_i$. Using Bernstein's inequality, for each column  of $F\odot \tl{F}$, we have, with probability $1-\delta$,
\[  \left| \,\| (F\odot \tl{F})_i\|_1 - |U|\cdot |T|\inner{\Ebb[\pi], (P)_i}\inner{\Ebb[\pi], (\tl{P})_i}\right| \leq \sqrt{ \frac{8 }{3} |U|\cdot |T|\cdot P_{\max}^4 \cdot \log \frac{|U|\cdot |T|}{\delta}},\]
by applying Bernstein's inequality, since  $\inner{\pi, (P)_i} \inner{\pi, (\tl{P})_i}\leq \max_i (P^\top \pi)_i (\tl{P}^\top \pi)_i \leq P_{\max}^2 ,$ 
 and   \begin{align*}&\max\left(\sum_{u\in U, t \in T}\|\Ebb[(P)_i^\top \pi_u \pi^\top_u (P)_i]
\cdot \Ebb[(\tl{P})_i^\top \pi_t \pi^\top_t (\tl{P})_i]\|, \sum_{u\in U,t\in T}\|\Ebb[ \pi_u^\top(P)_i(P)_i^\top \pi_u ]\cdot \Ebb[ \pi_t^\top(\tl{P})_i(\tl{P})_i^\top \pi_t ]\|\right) \\ &\leq |U|\cdot |T|\cdot P_{\max}^4 .\end{align*}
  The other results follow similarly.
\eprf\\

The lowest singular value for the Khatri-Rao product is a bit more involved and we provide the bound below.

\begin{lemma}[Spectral Bound for KR-product]\label{lemma:KR}
\[\sigma_k^2 ( F\odot \tl{F}) \geq |U|\cdot |T|\sigma_k(\Gamma*\Gamma) - \sqrt{ \frac{8}{3}  |U|\cdot |T|\cdot \|P\|^2\cdot \|\tl{P}\|^2\cdot \|\Ebb[\pi\pi^\top]\|^2 \cdot \log \frac{|U|\cdot |T|}{\delta}},\]
where $\Gamma:=P^\top \Ebb[\pi\pi^\top]\tl{P}$ and $*$ denotes Hadamard product.
\end{lemma}

\bprf The result in the Lemma follows directly from the concentration result. For the homogeneous setting, we have for a matrix $\Gamma$, \[\sigma_k(\Gamma*\Gamma) = \Theta\left( \min_i\Gamma(i,i)^2 -\max_{i\neq j} \Gamma(i,j)^2\right).\] Substituting we have the result. \eprf

\paragraph{Remark: }For the homogeneous setting, with $P=\tilde{P}$ having $p$ on the diagonal and $q$ on the off-diagonal, we have \begin{align*}\Gamma&= \left[(p-q)I + q 11^\top\right] \Ebb[\pi \pi^\top] \left[(p-q)I + q 11^\top\right]\\
&= (p-q)^2 \Ebb[\pi\pi^\top] + 2(p-q)q   v 1^\top + q^2 \|\Ebb[\pi \pi^\top]\|_{sum} 11^\top, \end{align*} where $v$ is a vector where $v_i = \|\Ebb[\pi \pi^\top]^{(i)}\|_1$, where $M^{(i)}$ denotes the $i^{\tha}$ row of $M$.
Thus, we have the following bound
\begin{align*}\sigma_k(\Gamma * \Gamma)&= \left( \min_{i, j\neq i}\left(\Gamma(i,i)^2 -  \Gamma(i,j)^2\right)\right)\\ &=\Theta\left( (p-q)^4\min_{i, j \neq i}\left( \Ebb(\pi_i^2)- \Ebb[\pi_i \pi_j]\right)^2\right),\end{align*}
assuming that $\Ebb[\pi_i^2]- \Ebb[\pi_i \pi_j] = \Theta(\Ebb[\pi_i^2])$ for all $i\neq j$, and the other terms which are dropped are positive.
Thus, we have w.h.p.
\beq \sigma_k(F\odot \tl{F}) = \Omega\left(n (p-q)^2\min_{i, j \neq i}\left(\Ebb[\pi_i^2] -\Ebb[\pi_i \pi_j]\right)\right)\eeq
%
%
%

%

\section{Standard Matrix Concentration and Perturbation Bounds}

\subsection{Bernstein's Inequalities}
One of the key tools we use is the standard matrix Bernstein inequality~\cite[thm. 6.1, 6.2]{tropp2012user}.

\begin{proposition}[Matrix Bernstein Inequality]\label{prop:bernstein}
Suppose $Z = \sum_j W_j$ where
\begin{enumerate}\itemsep 0pt
\item $W_j$ are independent random matrices with dimension $d_1\times d_2$,
\item $\E[W_j] = 0$ for all $j$,
\item $\norm{W_j} \le R$ almost surely.
\end{enumerate}

Let $d = d_1+d_2$, and $\sigma^2 = \max\left\{\norm{\sum_j\E[W_jW_j^\top ]},\norm{\sum_j\E[W_j^\top W_j]} \right\}$, then
we have

\begin{align*}
\Pr[\norm{Z} \ge t] \le& d \cdot exp\left\{\frac{-t^2/2}{\sigma^2 + Rt/3}\right\}\\ \le& d \cdot exp\left\{\frac{-3t^2}{8\sigma^2}\right\}, \quad t\leq \sigma^2/R, \\ \le& d \cdot exp\left\{\frac{-3t}{8R}\right\}, \quad t\geq \sigma^2/R
\end{align*}
\end{proposition}

\begin{proposition}[Vector Bernstein Inequality]\label{prop:vectorbernstein}
Let $z = (z_1, z_2, ..., z_n) \in \R^n$ be a random vector with independent entries, $\E[z_i] = 0$,
$\E[z_i^2] = \sigma_i^2$, and $\Pr[|z_i| \le 1] = 1$. Let $A = [a_1|a_2|\cdots|a_n] \in \R^{m\times n}$ be
a matrix, then

$$
\Pr[\norm{Az} \le (1+\sqrt{8t})\sqrt{\sum_{i=1}^n \norm{a_i}^2 \sigma_i^2} +(4/3)\max_{i\in[n]}
\norm{a_i} t] \ge 1-e^{-t}.
$$
\end{proposition}

\subsection{Hanson-Wright Inequalities}
We require the Hanson-Wright inequality~\cite{rudelson2013hanson}.

\begin{proposition}[Hanson-Wright Inequality: sub-Gaussian bound]\label{prop:hansonwright:subgaussian}
Let $z = (z_1, z_2, ..., z_n) \in \R^n$ be a random vector with independent entries, $\E[z_i] = 0$  and $\Pr[|z_i| \le 1] = 1$ and let $M\in \R^{n\times n}$ be any matrix. There exists a constant $c>0$ s.t.
\[\Pr\left[|z^\top M z - \Ebb(z^\top M z)|>t\right]\leq 2 \exp\left[-c \min\left(\frac{t^2}{\|M \|_{\Fbb}^2}, \frac{t}{\|M \|} \right)\right] \]
\end{proposition}

Unfortunately the sub-Gaussian bound is not strong enough when $z$ has small variance $\sigma^2$. In this case, we get the perturbation as $\tl{O}(\|M \|_{\Fbb})$ instead of $\tl{O}(\sigma\|M \|_{\Fbb})$, which is desired. This is because for a bounded random variable, the sub-Gaussian parameter only depends on the bound and not on the variance.

We will consider  an extension of the Hanson-Wright inequality to
 sub-exponential random variables~\cite{erdHos2010bulk,vu2013random} and employ the sub-exponential formulation for bounded random variables.
 We first define sub-exponential random variable~\cite[Definition 5.13]{vershynin2010introduction}.

 \begin{definition}[Sub-exponential Random Variable]\label{def:subexponential} A zero-mean random variable $X$ is said to be sub-exponential if there exists a parameter $K$ such that $ \Ebb[e^{X/K}] \leq e$.
  \end{definition}

\noindent{\em Remark: }There are other equivalent notions for sub-exponential random variables~\cite[Definition 5.13]{vershynin2010introduction}, but this will be the convenient one for proving sub-exponential bound for Bernoulli random variables.   It is easy to see that the centered Bernoulli random variables are sub-exponential for some constant $K$.

We will employ the following version of Hanson-Wright's inequality for sub-exponential random variables~\cite[Lemma B.2]{erdHos2010bulk}.

\begin{proposition}[Hanson-Wright Inequality: sub-exponential bound]\label{prop:hansonwright}
Let $z = (z_1, z_2, ..., z_n) \in \R^n$ be a random vector with independent entries, $\E[z_i] = 0$, $\Ebb[z_i^2]\leq\sigma^2$  and $z_i$ are sub-exponential  and let $M\in \R^{n\times n}$ be any matrix. There exists  constants $c,C>0$ s.t.
\[\Pr\left[|z^\top M z - \Ebb(z^\top M z)|>t\sigma^2\|M \|_{\Fbb}\right]\leq C \exp\left[-c  t^{1/4}\right]. \]
\end{proposition}

\noindent{\em Remark: }The result in the form above appears in~\cite[(13)]{vu2013random} and we set $\alpha=1$ in~\cite[(13)]{vu2013random}. The parameter $C$ above differs from the sub-exponential parameter $K$ by only a constant factor.


Comparing sub-exponential formulation in  Proposition~\ref{prop:hansonwright} with sub-Gaussian formulation in  Proposition~\ref{prop:hansonwright:subgaussian}, we see that in the former, the deviation is $\tl{O}(\|M\|_{\Fbb}\sigma)$, while in the latter it is only $\tl{O}(\|M\|_{\Fbb})$.
%
%

  %
%
%
%
%

Thus, for centered Bernoulli random variables  and we can employ Proposition~\ref{prop:hansonwright}, and we will use it for distance concentration bounds.

\subsection{Davis-Kahan Inequality}We also use the standard Davis and Kahan bound for subspace perturbation.


\begin{proposition}[Davis and Kahan]\label{prop:daviskahan}For a matrix $\hat{A}$, let   $\Projhat$ be the projection matrix on to  its top-$k$ left singular vectors. For any rank-$k$ matrix $A$, we have
\[\| (\Projhat- I)A\| \leq 2 \| \hat{A}-A\| \]
\end{proposition}

\bprf This is directly from \cite[Lemma 12]{McSherry01}. By writing $A= \hat{A} -( \hat{A}-A)$, we have
\[\| (\Projhat- I)A\| \leq  \| (\Projhat- I)\hat{A}\| +\| (\Projhat- I)( \hat{A}-A)\|   ,\] and each of the terms is less than $\| \hat{A}-A\|$. For the first term, it is because $\Projhat \hat{A}$ is the best rank-$k$ approximation of $\hat{A}$ and since $A$ is also rank $k$, the residual $\| (\Projhat- I)\hat{A}\| \leq \|\hat{A}-A\|$. For the second term, $\| (\Projhat- I)( \hat{A}-A)\|\leq  \|\hat{A}-A\|$ since  $(\Projhat- I)$ cannot increase norm.  \eprf

\end{document}